\documentclass[10pt,twocolumn,letterpaper]{article}

\usepackage[pagenumbers]{cvpr} 

\usepackage{bm}
\usepackage{mathtools}
\usepackage{amsthm}
\usepackage{multirow}
\usepackage{array}
\usepackage{adjustbox}
\usepackage{colortbl}
\usepackage{algorithm}
\usepackage[noend]{algpseudocode}
\algrenewcommand{\algorithmiccomment}[1]{\hfill{\color{algcmt}$\triangleright$~#1}}
\usepackage{microtype}

\definecolor{oursbg}{RGB}{232,241,250}
\definecolor{oursblue}{RGB}{0,114,178}
\definecolor{algcov}{HTML}{0072B2}
\definecolor{algref}{HTML}{2A8BC4}
\definecolor{alglgb}{HTML}{B07400}
\definecolor{alganc}{HTML}{009E73}
\definecolor{algveto}{HTML}{D55E00}
\definecolor{algcmt}{HTML}{555555}

\newcommand{\method}{AlignSpan\xspace}
\newcommand{\asWAB}{\textsc{wab}\xspace}
\newcommand{\asCVS}{\textsc{cvs}\xspace}
\newcommand{\best}[1]{{\boldmath\textbf{#1}}}
\newcommand{\second}[1]{\underline{#1}}
\newcommand{\na}{--}
\newcommand{\vloc}{\bm{v}^{\mathrm{loc}}}
\newcommand{\vglb}{\bm{v}^{\mathrm{glb}}}
\newcommand{\ceff}{c^{\mathrm{eff}}_t}

\usepackage{placeins}

\definecolor{cvprblue}{rgb}{0.21,0.49,0.74}
\usepackage[pagebackref,breaklinks,colorlinks,allcolors=cvprblue]{hyperref}

\usepackage[capitalize]{cleveref}
\crefname{section}{Sec.}{Secs.}
\Crefname{section}{Section}{Sections}
\Crefname{table}{Table}{Tables}
\crefname{table}{Tab.}{Tabs.}
\crefname{algorithm}{Alg.}{Algs.}
\crefname{proposition}{Prop.}{Props.}
\crefname{corollary}{Cor.}{Cors.}

\def\confName{CVPR}
\def\confYear{2027}

\title{Anisotropic Feature Caching for Training-Free Long Video Generation}

\author{
\bf Yuze Wu$^{\heartsuit}$,
Yuhan Chen$^{\heartsuit}$,
Sujio Cheng$^{\spadesuit}$,
Runmin Jiang$^{\bigstar}$,
Jiyue Jiang$^{\heartsuit}$$^{\bigstar}$\thanks{Corresponding Author and Project Leader.}\\
$^{\heartsuit}$ Department of Computer Science and Engineering, The Chinese University of Hong Kong,\\
$^{\spadesuit}$ Department of Computer Science and Information Systems, University of Limerick,\\
$^{\bigstar}$ School of Computer Science, Carnegie Mellon University\\
{\tt
\{1155244501, jiangjy\}@link.cuhk.edu.hk
}
}

\begin{document}
\maketitle
\begin{abstract}
Training-free long video generation commonly extends a short-clip diffusion transformer with dual-path sampling: at every denoising step a local path denoises overlapping temporal windows, a global path denoises temporally dilated clips, and the two predictions are fused. A 21\,s video then needs 11 to 14 backbone evaluations per step and several GPU-hours in total. Existing feature caches treat all of these evaluations alike, although their errors do not affect the output equally: the fusion weight scales the error of each path and drives the global contribution toward zero late in sampling, and only a non-overlapping subset of the local windows covers the sequence, while the remaining windows smooth the transitions between them. We present \method, a training-free cache scheduler that gives each path a reuse threshold inversely proportional to its fusion weight, reuses seam-refinement windows under a latent seam check, and prices every cache hit with an online-calibrated error model. On 40 VBench-Long prompts (21--23\,s videos), \method accelerates MiniMax-H3, HunyuanVideo, and Wan2.1-1.3B by $2.27\times$, $2.58\times$, and $2.13\times$, and on all three it is both faster than per-unit EasyCache and closer to the unaccelerated output ($+0.5$ to $+2.8$\,dB PSNR); a HunyuanVideo sample drops from 2.9\,h to 1.1\,h. Our code can be found in the following link: \url{https://github.com/jiangjyjy/AlignSpan}.
\end{abstract}

\section{Introduction}
\label{sec:intro}

\begin{figure}[t]
  \centering
  \includegraphics[width=\linewidth]{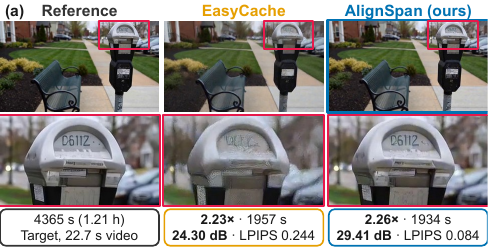}\\[2pt]
  \includegraphics[width=\linewidth]{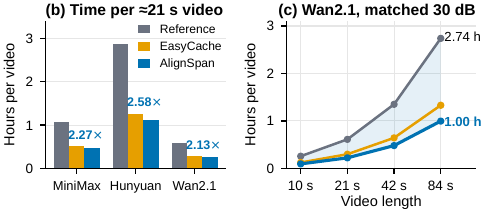}
  \vspace{-6pt}
  \caption{\textbf{Same cost, different fidelity.}
  (a)~A 22.7\,s MiniMax-H3 video from the 100-prompt VBench panel (\emph{``a parking meter on the right of a bench, front view''}, frame 28). At almost the same speed, EasyCache blurs the meter ID while \method keeps it legible and is $5.11$\,dB closer to the unaccelerated output.
  (b)~Mean generation time per video on one RTX~5090 for the 40-prompt VBench-Long panel (\cref{tab:main}).
  (c)~Wan2.1 time per video versus length at a matched 30\,dB (reference times from \cref{tab:cost}, speedups from \cref{tab:scaling}).}
  \label{fig:teaser}
\end{figure}

\begin{figure*}[t]
  \centering
  \includegraphics[width=\textwidth]{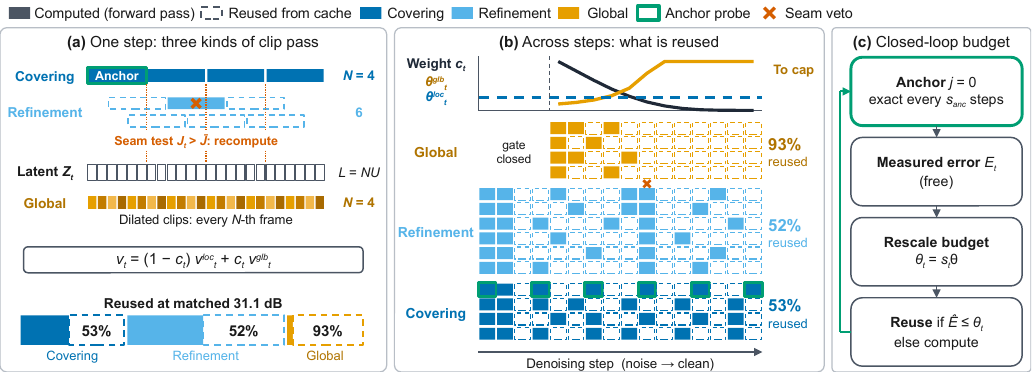}
  \vspace{-6pt}
  \caption{\textbf{Overview of \method.} Filled blocks are forward passes; dashed blocks are served from cache.
  (a)~One step with both paths active: $N$ covering windows tile the latent, $(m{-}1)(N{-}1)$ refinement windows smooth the boundaries between them, and $N$ dilated global clips enforce long-range consistency; the two paths are fused with weight $c_t$. Refinement windows are recomputed when the seam test $J_t$ fires. Bottom: fraction of each pass type served from cache at a matched 31.1\,dB (20.8\,s Wan2.1 videos, \cref{tab:axis}).
  (b)~Schematic schedule over the denoising steps. As $c_t\to0$ the global threshold grows as $1/c_t$ up to its cap, so global clips are reused most often (\asWAB); refinement windows are reused unless the seam test fires (\asCVS); covering windows are reused least.
  (c)~The anchor window, computed exactly every $s_{\mathrm{anc}}$ steps, provides a measurement of reuse error without an extra forward pass; it rescales the budget used by all reuse decisions.}
  \label{fig:overview}
  \vspace{-4pt}
\end{figure*}

Video diffusion transformers are trained on clips of a few seconds.
A common way to extend them beyond that horizon without retraining is \emph{dual-path} sampling~\cite{yang2026diffvf,qiu2024freenoise,lu2024freelong}.
At every denoising step, a \emph{local} path denoises overlapping temporal windows and blends them with triangular weights so that window boundaries do not show, while a \emph{global} path gathers temporally dilated clips that span the whole sequence, denoises them, and scatters the predictions back to keep the video consistent over its full length.
The two velocity fields are fused at each denoising step with a weight $c_t$ that decays over the course of sampling.

The construction is effective but expensive when generating longer videos.
Let the backbone process windows of $U$ latent frames and let the target video have $L{=}NU$ frames.
With window stride $s{=}U/m$ for an integer overlap factor $m$, a step in which both paths are active runs $m(N{-}1){+}1$ local windows and $N$ global clips, \ie $(m{+}1)N{-}(m{-}1)$ clip evaluations where a short video needs one, before classifier-free guidance doubles the count.
For Wan2.1 ($m{=}3$), videos of 21 to 84\,s ($N{=}4$ to $16$) therefore cost $14\times$ to $62\times$ a single clip per step (\cref{tab:cost}).
In wall-clock terms, one 21\,s HunyuanVideo sample takes 2.9\,h on an RTX~5090, or about eight GPU-hours per minute of output.

Feature caching is a natural remedy, and many methods accelerate diffusion transformers by reusing features across denoising steps~\cite{ma2024deepcache,zhao2024pab,liu2025teacache,lv2024fastercache,zhou2025easycache,selvaraju2024fora,chen2024deltadit}.
These methods were developed for single-clip sampling and exploit one kind of redundancy: the similarity of a clip's features at adjacent steps.
Applying such an adaptive cache independently to every window and clip is a strong baseline for the long-video setting, but it treats all evaluations in the dual-path loop as interchangeable.
They are not, and the differences follow directly from how the loop is built.

\noindent\textbf{The fusion weight acts as an error budget per path.}
Since the paths are combined as $(1{-}c_t)\vloc_t + c_t\vglb_t$, a caching error in the global path is scaled by $c_t$ before it reaches the solver, and an error in the local path by $1{-}c_t$.
Weight-Aware Budgeting (\asWAB) therefore gives each path a tolerance inversely proportional to its weight; as $c_t$ decays, the global path can be served almost entirely from cache in the second half of sampling.

\noindent\textbf{Only some local windows carry primary content.}
With stride $U/m$, the $N$ windows whose indices are multiples of $m$ tile $[0,L)$ without gaps or overlap.
The other $(m{-}1)(N{-}1)$ windows straddle the boundaries between them and only smooth the transitions, so an error in such a refinement window perturbs only the boundary region.
Covering-Set Skipping (\asCVS) reuses refinement windows more readily and recomputes them when a latent seam statistic indicates that a boundary is becoming visible.

To turn these observations into a scheduler, \method prices every candidate cache hit in the same unit, the relative velocity error, with an additive model calibrated online from the exact evaluations the sampler performs anyway.
A periodic exact evaluation of the first covering window measures the reuse error at no extra cost, and a feedback loop tightens the budget if this measurement exceeds the target.
\Cref{fig:teaser}a shows the effect on one video in a direct visual comparison: at the same speed as per-unit EasyCache, \method preserves fine detail that EasyCache blurs.

\noindent Our contributions are:
\begin{itemize}[leftmargin=*,itemsep=1pt,topsep=2pt]
\item A cost analysis of dual-path long-video sampling that splits each step into covering, refinement, and global passes, which differ systematically in how much reuse error they tolerate.
\item \method, a training-free cache scheduler that combines \asWAB, seam-gated \asCVS, and an online-calibrated error model with drift-corrected reuse and an anchor-window feedback loop.
\item An evaluation of fidelity to the unaccelerated output on three backbones and two prompt panels, with matched-fidelity sweeps: on VBench-Long \method reaches $2.27\times$, $2.58\times$, and $2.13\times$, faster and more faithful than per-unit EasyCache on each backbone.
\end{itemize}

\section{Related Work}
\label{sec:related}

\noindent\textbf{Training-free long video generation.}
Short-video backbones~\cite{wang2023lavie,kong2024hunyuanvideo,wan2025} have been extended without retraining through structured noise initialization~\cite{qiu2024freenoise,lu2024freelong,ge2023pyoco}, windowed and dilated denoising adapted from high-resolution image synthesis~\cite{bar2023multidiffusion,he2023scalecrafter}, positional-frequency scaling~\cite{zhao2025riflex}, and diagonal denoising queues~\cite{kim2024fifo,chen2025ouroboros}.
These extensions avoid even the parameter-efficient fine-tuning~\cite{wang2024prolora,wang2024lora,wang2025mos} that adapting a pretrained backbone would otherwise require.
We build on the dual-path construction that fuses overlapping local windows with dilated global clips~\cite{yang2026diffvf}.
The pipeline itself is left unchanged; our goal is to reduce its cost, which we analyze in \cref{sec:cost}.

\noindent\textbf{Feature caching for diffusion.}
Caching reuses intermediate features across denoising steps, either on a fixed schedule~\cite{ma2024deepcache,selvaraju2024fora,zhao2024pab,chen2024deltadit} or adaptively based on runtime change criteria~\cite{liu2025teacache,lv2024fastercache,zhou2025easycache,wimbauer2024blockcaching,kahatapitiya2025adacache}; TaylorSeer~\cite{liu2025taylorseer} extrapolates cached features with a Taylor expansion.
Other training-free accelerators include fast ODE solvers~\cite{lu2022dpmsolver,karras2022edm}, token merging and token-wise caching~\cite{bolya2023tome,zou2024tokencache}, and sparse or pruned attention~\cite{xi2025sparsevideogen,wang2024attendiff}, which are complementary to caching.
Training-side efficiency, such as data pruning for pretraining foundation models at scale~\cite{wu2026investigating}, is likewise orthogonal to inference-time acceleration.
Outside diffusion, reusing cached states has also been studied for the key-value caches of vision-language models~\cite{chen2026conduit}, where long multimodal contexts and multi-stage reasoning~\cite{zhou2025proreason} make recomputation costly.
To our knowledge, existing cache schedulers were designed for single-clip sampling and decide reuse along the step axis only.

\noindent\textbf{Calibrated cache scheduling.}
Adaptive caches usually decide reuse from a proxy such as the relative change of a feature, which can drift away from the true output error; recent work calibrates the decision online against measured errors~\cite{anon2026aligncache}.
In the multi-window loop such measurements come for free whenever a unit that holds a cache entry is computed exactly.

\section{Cost Structure of Dual-Path Sampling}
\label{sec:cost}

\noindent\textbf{Setup.}
Let the backbone have native window length $U$ latent frames and let the target sequence have $L{=}NU$ frames for integer extension factor $N$.
We write $\bm{Z}_t\in\mathbb{R}^{L\times C\times h\times w}$ for the long latent at denoising step $t\in\{T,\dots,1\}$.
The \emph{local path} (Weighted Window Sampling) extracts windows $\bm{Z}_t[js{:}js{+}U]$, $j=0,\dots,W_{\!L}{-}1$, at stride $s{=}U/m$, predicts per-window velocities $\vloc_{j,t}$, and blends them into $\vloc_t$ with normalized triangular weights that decay from window center to edge.
The \emph{global path} (Temporal Extended Sampling) gathers $N$ dilated clips $\bm{Z}_t[\iota{+}n\,{:}{:}\,N]$, predicts $\vglb_{n,t}$, and scatters them into $\vglb_t$; $\iota\in\{0,1\}$ keeps the single-frame first latent of causal 3D VAEs on the local path.
Stepping each window separately would desynchronize the history of multistep solvers such as UniPC, so the two paths are fused in velocity space and the solver is called once on the full sequence:
\begin{equation}
\begin{aligned}
\bm{v}_t&=(1-c_t)\,\vloc_t+c_t\,\vglb_t,\\
c_t&=\mathbb{I}[t<\tau T]\;\alpha\Bigl(\tfrac12\bigl(1+\cos\tfrac{\pi(T-t)}{T}\bigr)\Bigr)^{c_s},
\end{aligned}
\label{eq:fusion}
\end{equation}
where $\alpha$ is the initial global weight, $c_s$ shapes the decay, and $\tau\in(0,1]$ is a late-start gate: the global path is inactive while $t\ge\tau T$.

\noindent\textbf{Cost per step.}
With integer $m{=}U/s$ and $L{=}NU$ there are no trailing windows, so the window count is exact:
\begin{equation}
\begin{aligned}
W_{\!L}&=m(N{-}1){+}1=\underbrace{N}_{\text{covering}}+\underbrace{(m{-}1)(N{-}1)}_{\text{refinement}},\\
P_t&=W_{\!L}+g_tN,
\end{aligned}
\label{eq:cost}
\end{equation}
where $g_t=\mathbb{I}[c_t>c_{\mathrm{floor}}]$ indicates that the global path is active.
A step with an active global path (a \emph{gate-open} step) thus performs $P=(m{+}1)N{-}(m{-}1)$ clip evaluations, compared with one for a short video.
\Cref{eq:cost} distinguishes three kinds of pass:
\textbf{covering windows} $j\in\mathcal{C}=\{0,m,\dots,(N{-}1)m\}$, which tile $[0,L)$ with no gaps or mutual overlap, so every frame depends on its covering window;
\textbf{refinement windows} $j\notin\mathcal{C}$, which straddle the boundaries between covering windows and smooth the transitions between adjacent covering windows; and
\textbf{global clips}, which enforce long-range consistency and enter the output scaled by $c_t$.
On Wan2.1 at $N{=}4$, $m{=}3$, a gate-open step runs $4{+}6{+}4{=}14$ passes, and over a 50-step run (37 gate-open steps, 1296 CFG forwards) the realized shares are $30.9/46.3/22.8\%$ (\cref{tab:cost}).
About $69\%$ of all backbone work is therefore spent either on seam refinement or on global passes whose weight decays toward zero.

\begin{table}[t]
\centering
\caption{\textbf{Cost of dual-path sampling} (Wan2.1-1.3B, 16\,fps, $m{=}3$, unaccelerated, 10 prompts): clip passes per gate-open step from \cref{eq:cost} (the total is also the ratio to a short clip), mean wall-clock of a gate-open step, and end-to-end time per video. CFG doubles all raw forwards and is omitted. At $N{=}4$ the covering/refinement/global pass shares are $28.6/42.9/28.6\%$ of a gate-open step and $30.9/46.3/22.8\%$ ($400/600/296$ forwards) of a full 50-step run.}
\label{tab:cost}
\small
\setlength{\tabcolsep}{3.3pt}
\begin{tabular}{lccccccc}
\toprule
$N$ & Length & Cov. & Ref. & Glb. & Total & Step (s) & Video \\
\midrule
1 & 5.1\,s & 1 & 0 & 0 & 1 & 3.4 & 3.0\,min \\
2 & 10.3\,s & 2 & 2 & 2 & 6 & 20.0 & 15.4\,min \\
4 & 20.8\,s & 4 & 6 & 4 & 14 & 47.1 & 36.7\,min \\
8 & 41.8\,s & 8 & 14 & 8 & 30 & 103.9 & 1.35\,h \\
16 & 83.8\,s & 16 & 30 & 16 & 62 & 209.8 & \textbf{2.74\,h} \\
\bottomrule
\end{tabular}
\end{table}

\noindent\textbf{Where reuse is possible.}
Let $\bm{v}_{u,t}$ be the post-guidance velocity of unit $u$ (a covering window, a refinement window, or a global clip) at step $t$.
Existing caches exploit the \emph{step axis}: $\bm{v}_{u,t}\approx\bm{v}_{u,t'}$ for nearby steps.
The dual-path loop adds a second, \emph{structural} axis: the weight $c_t$ scales each path's contribution to the output error (\cref{sec:wab}), and the covering/refinement split separates primary denoising from boundary smoothing (\cref{sec:cvs}).
\method uses both.

\section{Method: \method}
\label{sec:method}

\Cref{fig:overview} summarizes the method.
All thresholds, probe readings, and estimator entries in \method are expressed in one dimensionless quantity, the \emph{relative RMS velocity error}
\begin{equation}
e(\hat{\bm{v}},\bm{v})=\frac{\operatorname{rms}(\hat{\bm{v}}-\bm{v})}{\operatorname{rms}(\bm{v})+\epsilon},\quad \operatorname{rms}(\bm{x})=\sqrt{\tfrac1d\lVert\bm{x}\rVert_2^2},
\label{eq:unit}
\end{equation}
computed in single precision on post-guidance velocities.
When unit $u$ is served from its own entry $(\bm{v}^{\mathrm{c}}_u,\bm{z}^{\mathrm{c}}_u)$ cached at an earlier step, \method applies a first-order input-drift correction,
\begin{equation}
\hat{\bm{v}}_{u,t}=\bm{v}^{\mathrm{c}}_u+\bigl(\bm{z}_{u,t}-\bm{z}^{\mathrm{c}}_u\bigr),
\label{eq:drift}
\end{equation}
which corresponds to the output-minus-input residual reuse of adaptive step caches~\cite{zhou2025easycache} and lets chains of reuse follow the drifting input.
All error observations evaluate \cref{eq:unit} on this corrected $\hat{\bm{v}}_{u,t}$, so the error model prices exactly the tensor that is passed to the solver.

\subsection{Weight-Aware Budgeting (\asWAB)}
\label{sec:wab}

Suppose caching introduces velocity errors $\bm{\delta}^{\mathrm{loc}}_t$ and $\bm{\delta}^{\mathrm{glb}}_t$ in the two paths.
By \cref{eq:fusion} and the triangle inequality, the error of the fused velocity satisfies
\begin{equation}
\lVert\bm{\delta}_t\rVert\le(1-\ceff)\lVert\bm{\delta}^{\mathrm{loc}}_t\rVert+\ceff\lVert\bm{\delta}^{\mathrm{glb}}_t\rVert,
\label{eq:budget}
\end{equation}
with effective weight $\ceff=c_t\,\mathbb{I}[c_t>c_{\mathrm{floor}}]$.
Given a step budget $\theta_t=s_t\theta$, where $s_t$ is the feedback scale of \cref{sec:probe}, assigning half of the budget to each path gives asymmetric thresholds, capped at $\theta_{\max}$:
\begin{equation}
\theta^{\mathrm{loc}}_t=\min\Bigl(\theta_{\max},\frac{\theta_t}{2(1-\ceff)}\Bigr),\quad
\theta^{\mathrm{glb}}_t=\min\Bigl(\theta_{\max},\frac{\theta_t}{2\ceff}\Bigr).
\label{eq:wab}
\end{equation}
While the global path is off ($\ceff{=}0$), global clips are not requested and do not count toward any reuse rate.
A single threshold shared by both paths, as in a short-video cache applied per unit, implicitly treats the paths as equally weighted ($c_t{=}\tfrac12$), whereas the schedule starts at $\alpha\le0.5$ and decays quickly.
As $c_t\to0$, $\theta^{\mathrm{glb}}_t$ grows to its cap late in sampling, when the coarse layout has formed and global velocities change slowly.
At a matched 31.1\,dB on Wan2.1, \method reuses $92.6\%$ of the active global passes (\cref{tab:axis}).
\Cref{eq:budget} bounds the error of the aggregated paths, while \cref{eq:wab} is applied to individual units in the relative metric of \cref{eq:unit}.
We therefore use \cref{eq:wab} for the \emph{ratio} of the two thresholds, $\theta^{\mathrm{glb}}_t/\theta^{\mathrm{loc}}_t=(1{-}\ceff)/\ceff$, and leave their overall scale to the budget $\theta$ and the feedback loop.

\subsection{Covering-Set Skipping (\asCVS)}
\label{sec:cvs}

The covering set $\mathcal{C}=\{j:j\equiv0\bmod m\}$ tiles $[0,L)$ exactly, and the refinement windows $\mathcal{R}=\{j:j\not\equiv0\bmod m\}$ straddle the boundaries $b\in\{U,2U,\dots,(N{-}1)U\}$.
Let $\mathcal{S}$ index the frames within radius $r{=}1$ of a boundary and $\mathcal{I}$ the remaining interior frames.
With the temporal second difference $\nabla^2_t\bm{z}_i=\bm{z}_{i+1}-2\bm{z}_i+\bm{z}_{i-1}$, we define the seam ratio
\begin{equation}
J_t=\frac{\mathbb{E}_{i\in\mathcal{S}}\lVert\nabla^2_t\bm{Z}_{t-1}[i]\rVert^2}{\mathbb{E}_{i\in\mathcal{I}}\lVert\nabla^2_t\bm{Z}_{t-1}[i]\rVert^2},
\label{eq:seam}
\end{equation}
which normalizes boundary roughness by the roughness of the same latent away from the boundaries.
Rather than dropping refinement windows, which would change the blending profile, \method keeps every window in the blend and decides per window whether to compute or reuse its velocity.
If $J_{t+1}$ exceeds the gate $\bar{J}_{t+1}$ ($j_{\mathrm{margin}}J^{\mathrm{ref}}_{t+1}$ when a reference seam trace is available, $J_{\max}$ otherwise), all refinement windows are recomputed, which also refreshes their cache entries and error statistics.
Otherwise refinement window $j$ is served its drift-corrected velocity (\cref{eq:drift}) whenever its estimated error satisfies $\hat{E}(\text{ref},a,\kappa)\le\theta^{\mathrm{loc}}_t$.
Seam smoothing matters most late in sampling, where $1{-}c_t\to1$, whereas the global path is active early; \asCVS and \asWAB therefore save compute in different phases of sampling.

\subsection{Error Model and Feedback}
\label{sec:probe}

\noindent\textbf{Additive error estimate.}
A candidate cache hit that serves a unit of kind $k\in\{\text{cov},\text{ref},\text{glb}\}$ from an entry of age $a$ (in steps) during timestep bin $\kappa\in\{5,15,\dots,45\}$ is priced as
\begin{equation}
\hat{E}=e_{\mathrm{step}}(k,a,\kappa)+\eta,
\label{eq:transfer}
\end{equation}
with a safety margin $\eta$ and $e_{\mathrm{step}}{=}0$ for $a{=}0$.
Each entry of $e_{\mathrm{step}}$ is a Welford accumulator that returns the upper estimate $\mu+\lambda_{\mathrm{std}}\sigma$ once it holds $n_{\min}$ observations.
It is updated whenever a unit that holds a cache entry of age $a\ge1$ is computed exactly, since the cached and the exact velocity are then both available.
An age without observations is priced from the largest calibrated lower age $a'$ as $\operatorname{upper}(k,a',\kappa)\sqrt{a/a'}$; if none exists, the unit is computed.
All tables are reset between samples, so the estimate adapts to each prompt.

\noindent\textbf{Anchor probe.}
Covering window $j{=}0$ is computed exactly every $s_{\mathrm{anc}}\ge1$ steps.
Whenever this happens while it holds a cached entry of age $a\ge1$, we record
\begin{equation}
E_t=e\bigl(\hat{\bm{v}}_{0,t},\bm{v}_{0,t}\bigr),
\label{eq:probe}
\end{equation}
with $\hat{\bm{v}}_{0,t}$ from \cref{eq:drift}, \ie the error that reuse would have caused at this step.
A stride $s_{\mathrm{anc}}{>}1$ keeps the anchor cheap at aggressive budgets and still yields $\lceil T/s_{\mathrm{anc}}\rceil$ readings per sample.

\noindent\textbf{Feedback on the budget.}
On probe steps with index $i$ in $i/(T{-}1)\in[0.1,0.9]$, the budget scale $s_t\in[s_{\min},1]$ is updated multiplicatively,
\begin{equation}
s_{t-1}=\operatorname{clip}\Bigl(s_t\exp\bigl(\beta\operatorname{clip}(\log\tfrac{\theta}{E_t},\pm c_{\mathrm{clip}})\bigr),s_{\min},1\Bigr),
\label{eq:controller}
\end{equation}
which tightens the thresholds when the measured error exceeds the budget and relaxes them otherwise, up to the nominal budget.
Comparing $E_t$ with $\theta$ rather than with $\hat{E}$ prevents the conservative upper estimate from relaxing the thresholds without bound.

\subsection{Algorithm and Overhead}
\label{sec:alg}

\begin{algorithm}[t]
\caption{One denoising step of \method.}
\label{alg:main}
\small
\begin{algorithmic}[1]
\Require $\bm{Z}_t$, $c_t$, budget $\theta$, scale $s_t$, seam ratio $J_{t+1}$, step index $i$
\State $\ceff\gets c_t\mathbb{I}[c_t{>}c_{\mathrm{floor}}]$;\; {\color{alglgb}$\theta^{\mathrm{loc}}_t,\theta^{\mathrm{glb}}_t\gets$ \cref{eq:wab}} \Comment{{\color{alglgb}\asWAB}}
\For{{\color{algcov}covering windows} $j\in\mathcal{C}$}
  \If{$j{=}0$ \textbf{and} $i\bmod s_{\mathrm{anc}}{=}0$} {\color{alganc}compute; record $E_t$}
  \ElsIf{$\hat{E}(\text{cov},a,\kappa)\le\theta^{\mathrm{loc}}_t$} serve $\hat{\bm{v}}_{j,t}$ (\cref{eq:drift})
  \Else{} compute; update $e_{\mathrm{step}}$
  \EndIf
\EndFor
\For{{\color{algref}refinement windows} $j\in\mathcal{R}$} \Comment{{\color{algref}\asCVS}}
  \If{{\color{algveto}$J_{t+1}\le\bar{J}_{t+1}$} \textbf{and} $\hat{E}(\text{ref},a,\kappa)\le\theta^{\mathrm{loc}}_t$} serve $\hat{\bm{v}}_{j,t}$
  \Else{} compute; update $e_{\mathrm{step}}$
  \EndIf
\EndFor
\If{$\ceff>0$} \Comment{{\color{alglgb}global path}}
  \For{{\color{alglgb}global clips} $n$}
    \If{$\hat{E}(\text{glb},a,\kappa)\le{\color{alglgb}\theta^{\mathrm{glb}}_t}$} serve $\hat{\bm{v}}_{n,t}$
    \Else{} compute; update $e_{\mathrm{step}}$
    \EndIf
  \EndFor
\EndIf
\State blend $\vloc_t$, scatter $\vglb_t$, fuse $\bm{v}_t$ (\cref{eq:fusion}); $\bm{Z}_{t-1}\gets\mathrm{Solver}(\bm{Z}_t,\bm{v}_t)$
\State evaluate $J_t$ on $\bm{Z}_{t-1}$ (\cref{eq:seam}); if a probe was recorded, {\color{alganc}update $s_{t-1}$} (\cref{eq:controller})
\end{algorithmic}
\end{algorithm}

\Cref{alg:main} lists one step.
All scheduling operations are reductions over clip tensors already on the GPU. On VBench-Long the measured overhead of \method is $0.15$--$1.14\%$ of forward time, within the run-to-run timing variation of the non-caching baselines ($-0.22$ to $+2.20\%$, \cref{tab:main}).
The cache holds the latest velocity and input clip of each unit, at most $(m{+}1)N{-}(m{-}1)$ clip-sized tensors, and raises peak memory by $0.9$--$2.0\%$ over the unaccelerated pipeline, less than per-unit EasyCache on all three backbones (\cref{tab:main}).

\section{Experiments}
\label{sec:experiments}

\begin{table*}[t]
\centering
\caption{\textbf{Main results on VBench-Long} (40 prompts, $N{=}4$, seed 0, one RTX~5090 per video): 22.7\,s (MiniMax-H3), 21.2\,s (HunyuanVideo), and 20.8\,s (Wan2.1) videos. Fidelity is measured against the unaccelerated dual-path output of the same pipeline. Best and second-best accelerators in efficiency and fidelity are in \best{bold} and \second{underline}; VBench-Long and resource columns are not ranked. Overhead is $(t_{\mathrm{den}}-\mathrm{NFE}\cdot t_{\mathrm{fwd}})/(\mathrm{NFE}\cdot t_{\mathrm{fwd}})$ with denoising time $t_{\mathrm{den}}$ and the reference's time per forward $t_{\mathrm{fwd}}$. $^\dagger$Re-run in a later, slower session (reference not re-run); corrected for the per-forward slowdown: $2.22\times$, $+0.70\%$. Gray row: budget $\theta{=}0.24$ instead of $0.21$, not ranked.}
\label{tab:main}
\small
\setlength{\tabcolsep}{3.35pt}
\begin{adjustbox}{max width=\textwidth}
\begin{tabular}{l cc ccc ccc cc}
\toprule
\multirow{2}{*}{\textbf{Method}} & \multicolumn{2}{c}{\textbf{Efficiency}} & \multicolumn{3}{c}{\textbf{Fidelity to reference}} & \multicolumn{3}{c}{\textbf{VBench-Long (not ranked)}} & \multicolumn{2}{c}{\textbf{Resource (not ranked)}} \\
\cmidrule(lr){2-3}\cmidrule(lr){4-6}\cmidrule(lr){7-9}\cmidrule(lr){10-11}
& Lat.\,(s)$\downarrow$ & Speedup$\uparrow$ & PSNR$\uparrow$ & SSIM$\uparrow$ & LPIPS$\downarrow$ & Temporal & Frame & Text & Peak (GiB) & Overhead (\%) \\
\midrule
\multicolumn{11}{l}{\emph{MiniMax-H3: 22.7\,s videos (544 frames, 24\,fps, $960{\times}544$)}}\\
Unaccelerated reference & 3843.2 & $1.00\times$ & \na & \na & \na & 0.9263 & 0.6266 & 0.3016 & 16.27 & \na \\
Reduced steps ($T'{=}22$) & \second{1727.3} & \second{$2.23\times$} & 18.11 & 0.641 & 0.316 & 0.9220 & 0.6278 & 0.3022 & 16.27 & $+1.38$ \\
Uniform interval ($q{=}2$) & 2099.8 & $1.83\times$ & 18.73 & 0.661 & 0.285 & 0.9212 & 0.6273 & 0.3023 & 16.35 & $+2.20$ \\
Per-unit adaptive (EasyCache)$^\dagger$ & 1846.7 & $2.08\times$ & \second{22.92} & \second{0.707} & \second{0.274} & 0.9047 & 0.5930 & 0.2949 & 16.60 & $+7.37$ \\
\rowcolor{oursbg}\textbf{\method (ours)} & \best{1690.6} & \best{$2.27\times$} & \best{25.68} & \best{0.781} & \best{0.181} & 0.9143 & 0.6149 & 0.2991 & 16.41 & $+0.30$ \\
\midrule
\multicolumn{11}{l}{\emph{HunyuanVideo: 21.2\,s videos (509 frames, 24\,fps, $960{\times}544$)}}\\
Unaccelerated reference & 10381.4 & $1.00\times$ & \na & \na & \na & 0.9023 & 0.6108 & 0.3069 & 23.72 & \na \\
Reduced steps ($T'{=}22$) & 4597.6 & $2.26\times$ & 17.71 & 0.690 & 0.348 & 0.8993 & 0.5716 & 0.3027 & 23.72 & $-0.22$ \\
Uniform interval ($q{=}2$) & 5482.1 & $1.89\times$ & 19.23 & 0.730 & 0.286 & 0.8951 & 0.5923 & 0.3045 & 23.91 & $-0.22$ \\
Per-unit adaptive (EasyCache) & \second{4558.0} & \second{$2.28\times$} & \second{30.17} & \second{0.874} & \second{0.092} & 0.8871 & 0.6053 & 0.3058 & 24.45 & $-0.28$ \\
\rowcolor{oursbg}\textbf{\method (ours)} & \best{4021.5} & \best{$2.58\times$} & \best{30.65} & \best{0.894} & \best{0.075} & 0.8904 & 0.6065 & 0.3063 & 23.97 & $+0.15$ \\
\midrule
\multicolumn{11}{l}{\emph{Wan2.1-1.3B: 20.8\,s videos (333 frames, 16\,fps, $832{\times}480$)}}\\
Unaccelerated reference & 2082.4 & $1.00\times$ & \na & \na & \na & 0.8979 & 0.6221 & 0.2977 & 8.02 & \na \\
Reduced steps ($T'{=}22$) & \best{940.8} & \best{$2.21\times$} & 14.50 & 0.547 & 0.434 & 0.8873 & 0.5822 & 0.2787 & 8.02 & $+1.57$ \\
Uniform interval ($q{=}2$) & 1121.3 & $1.86\times$ & 15.03 & 0.566 & 0.414 & 0.8847 & 0.5828 & 0.2821 & 8.24 & $+1.53$ \\
Per-unit adaptive (EasyCache) & 1018.3 & $2.05\times$ & \second{28.80} & \second{0.880} & \second{0.059} & 0.8916 & 0.6182 & 0.2977 & 8.45 & $+1.43$ \\
\rowcolor{oursbg}\textbf{\method (ours, $\theta{=}0.21$)} & \second{977.2} & \second{$2.13\times$} & \best{31.36} & \best{0.913} & \best{0.037} & 0.8950 & 0.6198 & 0.2977 & 8.18 & $+1.14$ \\
\color{gray}\method ($\theta{=}0.24$) & \color{gray}875.9 & \color{gray}$2.38\times$ & \color{gray}30.03 & \color{gray}0.897 & \color{gray}0.047 & \color{gray}0.8936 & \color{gray}0.6191 & \color{gray}0.2975 & \color{gray}8.18 & \color{gray}$+2.04$ \\
\bottomrule
\end{tabular}
\end{adjustbox}
\end{table*}

\begin{table*}[t]
\centering
\caption{\textbf{Results on a 100-prompt VBench panel} ($N{=}4$, same lengths and settings as \cref{tab:main}, separate campaign). VBench is the total score (reference: 0.8222, 0.7776, 0.7685), not ranked. Full table: supplementary \cref{tab:vb100full}.}
\label{tab:vbench100}
\small
\setlength{\tabcolsep}{3.0pt}
\begin{adjustbox}{max width=\textwidth}
\begin{tabular}{l ccccc ccccc ccccc}
\toprule
& \multicolumn{5}{c}{\textbf{MiniMax-H3} (22.7\,s)} & \multicolumn{5}{c}{\textbf{HunyuanVideo} (21.2\,s)} & \multicolumn{5}{c}{\textbf{Wan2.1-1.3B} (20.8\,s)} \\
\cmidrule(lr){2-6}\cmidrule(lr){7-11}\cmidrule(lr){12-16}
\textbf{Method} & Speedup$\uparrow$ & PSNR$\uparrow$ & SSIM$\uparrow$ & LPIPS$\downarrow$ & VBench & Speedup$\uparrow$ & PSNR$\uparrow$ & SSIM$\uparrow$ & LPIPS$\downarrow$ & VBench & Speedup$\uparrow$ & PSNR$\uparrow$ & SSIM$\uparrow$ & LPIPS$\downarrow$ & VBench \\
\midrule
Reduced steps & $2.25\times$ & 18.78 & 0.705 & 0.310 & 0.8176 & $2.26\times$ & 20.48 & 0.783 & 0.313 & 0.7624 & \second{$2.26\times$} & 16.30 & 0.612 & 0.396 & 0.7464 \\
Uniform interval & $1.85\times$ & 19.41 & 0.719 & 0.285 & 0.8200 & $1.89\times$ & 22.56 & 0.823 & 0.240 & 0.7653 & $1.89\times$ & 16.84 & 0.631 & 0.376 & 0.7525 \\
EasyCache & \second{$2.26\times$} & \second{24.65} & \second{0.772} & \second{0.234} & 0.7915 & \second{$2.37\times$} & \second{31.78} & \second{0.905} & \second{0.111} & 0.7629 & $2.16\times$ & \second{29.62} & \second{0.886} & \second{0.059} & 0.7625 \\
\rowcolor{oursbg}\textbf{\method} & \best{$2.53\times$} & \best{26.70} & \best{0.814} & \best{0.181} & 0.8007 & \best{$2.69\times$} & \best{32.57} & \best{0.925} & \best{0.078} & 0.7674 & \best{$2.28\times$} & \best{32.01} & \best{0.916} & \best{0.039} & 0.7660 \\
\bottomrule
\end{tabular}
\end{adjustbox}
\end{table*}

\begin{figure*}[t]
  \centering
  \includegraphics[width=\textwidth]{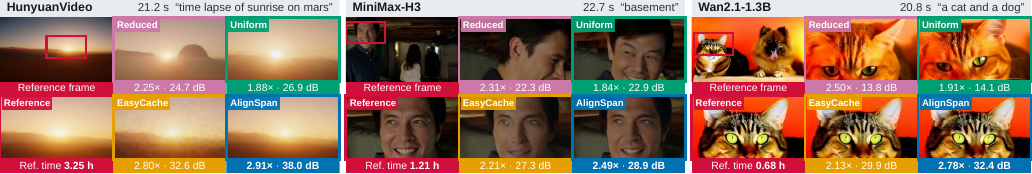}
  \vspace{-6pt}
  \caption{\textbf{Case studies} from the 100-prompt VBench panel ($N{=}4$, 20.8--22.7\,s videos). Each block shows a reference frame, its boxed zoom with the reference generation time, and the same zoom for each accelerator with the speedup and PSNR of \emph{this} video; the best speedup and PSNR per block are in bold. Reduced steps and uniform caching change the content (a dome on Mars, a different face, a different cat); EasyCache adds grain and blur. More cases: supplementary Figs.~S3--S10.}
  \label{fig:case}
  \vspace{-4pt}
\end{figure*}

\noindent\textbf{Reference and metrics.}
Unlike generation quality, acceleration has an exact reference: the unaccelerated dual-path output for the same prompt, seed, and schedule.
We decode both videos and report PSNR, SSIM, and LPIPS against this reference, so a higher PSNR means a more faithful acceleration, not a better generator.
We also report reference-free VBench scores as a sanity check on overall generation quality but do not rank by them, since an output that departs from the reference can still score well.

\noindent\textbf{Prompt sets.}
The main panel uses 40 long, descriptive prompts (48 words on average) scored with the VBench-Long protocol~\cite{huang2025vbenchpp}, which cuts each long video into short clips (11--13 per video here) and averages the VBench dimensions over clips.
We group its eight dimensions into Temporal (subject and background consistency, motion smoothness, dynamic degree), Frame (aesthetic and imaging quality), and Text (overall consistency, CLIP score); see \cref{eq:vblgroups} in the supplement for the formulas.
A second panel uses a random subset of 100 prompts from the VBench prompt suite~\cite{huang2024vbench} and the VBench quality, semantic, and total scores.
Sweeps and ablations (\cref{tab:isoquality,tab:axis,tab:scaling,tab:ablation}) use a fixed 10-prompt Wan2.1 suite, selected by the NFE difference between \method and EasyCache in a pilot run. All videos use seed 0 on one RTX~5090.

\noindent\textbf{Backbones.}
We use three diffusion transformers with joint spatio-temporal attention in the dual-path pipeline of~\cite{yang2026diffvf}:
\textbf{MiniMax-H3} ($960{\times}544$, 24\,fps, $U{=}40$, $m{=}2$),
\textbf{HunyuanVideo}~\cite{kong2024hunyuanvideo} ($960{\times}544$, 24\,fps, $U{=}32$, $m{=}2$), and
\textbf{Wan2.1-1.3B}~\cite{wan2025} ($832{\times}480$, 16\,fps, $U{=}21$, $m{=}3$).
Main comparisons use $N{=}4$ (20.8--22.7\,s) and 50 steps; all rows of a backbone differ only in the caching policy.

\noindent\textbf{Baselines.}
(1)~\textbf{Reduced steps} lowers the step count to $T'{=}22$ to match the accelerated latency;
(2)~\textbf{Uniform interval} recomputes every unit every $q{=}2$ steps;
(3)~\textbf{Per-unit adaptive} applies EasyCache~\cite{zhou2025easycache} independently to every local window and global clip;
(4)~\textbf{Path-uniform} is \method with a single threshold for both paths (\asWAB off); and
(5)~\textbf{Step-only} restricts \method to reuse along the step axis.
\method uses $\theta{=}0.21$ and anchor stride $s_{\mathrm{anc}}{=}3$ ($\theta{=}0.26$ and $s_{\mathrm{anc}}{=}7$ on MiniMax-H3).
EasyCache uses threshold $1.0$ and warm-up $R{=}15$ on MiniMax-H3, and $0.05$ and $R{=}10$ elsewhere.

\subsection{Main Results}
\label{sec:main}

\noindent\textbf{VBench-Long.}
On all three backbones \method is the most faithful accelerator and is faster than per-unit EasyCache (\cref{tab:main}).
Against EasyCache it gains $2.77$, $0.48$, and $2.56$\,dB PSNR and lowers LPIPS by $34\%$, $19\%$, and $36\%$ on MiniMax-H3, HunyuanVideo, and Wan2.1.
On HunyuanVideo it is also $13\%$ faster ($2.58\times$ vs.\ $2.28\times$).
On MiniMax-H3 the EasyCache run was repeated in a slower session; after correcting for this, EasyCache reaches $2.22\times$ against $2.27\times$, so on this backbone the two methods cost about the same and differ mainly in fidelity.
On Wan2.1 reduced steps is $4\%$ faster than \method at the default budget, but only at $14.5$\,dB.
Raising the budget to $\theta{=}0.24$ gives $2.38\times$, faster than every baseline, at $30.0$\,dB, still $1.2$\,dB above EasyCache.
Reduced steps and uniform caching stay below $20$\,dB on every backbone.

The VBench-Long group scores of \method stay within $0.012$ of the reference on every backbone, against up to $0.034$ for EasyCache and $0.040$ for reduced steps, and on 8 of the 9 backbone--group pairs they are closer to the reference than those of EasyCache (the ninth is a tie).
The scores also show why they are not used for ranking: reduced steps on MiniMax-H3 lies within $0.005$ of the reference on all three groups at only $18.1$\,dB.
Among the individual dimensions (supplementary \cref{tab:vbldims}), dynamic degree drops most under caching, and less for \method than for EasyCache on every backbone ($0.585$ vs.\ $0.544$, $0.389$ vs.\ $0.371$, and $0.459$ vs.\ $0.436$; reference $0.652$, $0.468$, $0.480$).

\noindent\textbf{A second prompt panel.}
The 100-prompt VBench panel (\cref{tab:vbench100}) gives the same ordering.
\method reaches $2.53\times$, $2.69\times$, and $2.28\times$, the highest speedup on all three backbones, against $2.26\times$, $2.37\times$, and $2.16\times$ for EasyCache, while gaining $0.8$ to $2.4$\,dB PSNR.
The VBench total score of every accelerator stays within $0.031$ of the reference.

\noindent\textbf{Generation time.}
A 21\,s HunyuanVideo sample drops from $2.88$\,h to $1.12$\,h ($1.27$\,h with EasyCache; supp.\ \cref{tab:time}). An 84\,s Wan2.1 video drops from $2.74$\,h to about one hour at 30\,dB (\cref{fig:teaser}c).

\noindent\textbf{Failure modes of the baselines.}
The case studies (\cref{fig:case}) show two kinds of failure.
Reduced steps and uniform caching change the \emph{content}: a stop sign reads ``MICITN'', a smiling man turns away from the camera, a tabby cat becomes a ginger one.
Per-unit EasyCache keeps the content but loses \emph{detail}: grain appears in flat regions and thin structures such as text and hair are smeared, consistent with reusing covering windows as readily as refinement windows and global clips.
Across all case studies, \method has the highest PSNR in every case and the highest speed in nearly all (supp.\ \cref{tab:cases}).

\subsection{Matched Fidelity and Attribution}
\label{sec:attr}

\begin{table}[t]
\centering
\caption{\textbf{Speedup at matched fidelity} on Wan2.1-1.3B at $N{=}2$, 10.3\,s videos ($n{=}10$). Each method's budget is swept over at least eight values and the end-to-end speedup is interpolated at fixed PSNR. AUC integrates speedup over $[28,34]$\,dB divided by 6\,dB. Uniform interval never reaches the band; $^\dagger$ marks its $q{=}2$ point (17.66\,dB).}
\label{tab:isoquality}
\small
\setlength{\tabcolsep}{4.9pt}
\begin{tabular}{lccccc}
\toprule
\multirow{2}{*}{\textbf{Method}} & \multicolumn{4}{c}{\textbf{Speedup at PSNR}$\uparrow$} & \multirow{2}{*}{\textbf{AUC}$\uparrow$} \\
\cmidrule(lr){2-5}
& 28\,dB & 30\,dB & 32\,dB & 34\,dB & \\
\midrule
Reduced steps & 1.00 & 0.99 & 1.00 & 1.00 & 1.00 \\
Uniform ($q{=}2$) & $1.89^\dagger$ & \na & \na & \na & \na \\
Per-unit adaptive & 2.46 & 2.08 & 1.77 & 1.43 & 1.93 \\
Path-uniform (abl.) & \second{2.71} & \second{2.30} & \second{1.90} & \second{1.57} & \second{2.11} \\
Step-only (abl.) & 1.73 & 1.66 & 1.53 & 1.33 & 1.57 \\
\rowcolor{oursbg}\textbf{\method} & \best{3.00} & \best{2.73} & \best{2.27} & \best{1.84} & \best{2.47} \\
\bottomrule
\end{tabular}
\end{table}

\begin{table}[t]
\centering
\caption{\textbf{Attribution to reuse axes} on Wan2.1, $N{=}4$, 20.8\,s videos, NFE speedup interpolated at 31.1\,dB ($n{=}10$). Reuse rates are over active requests. \asWAB has no row of its own because it only redistributes the budget over candidates supplied by the other axes.}
\label{tab:axis}
\small
\setlength{\tabcolsep}{3.6pt}
\begin{tabular}{lcccc}
\toprule
\multirow{2}{*}{\textbf{Axes used}} & \multirow{2}{*}{\textbf{Speedup}$\uparrow$} & \multicolumn{3}{c}{\textbf{Reused (\%)}} \\
\cmidrule(lr){3-5}
& & Cov. & Ref. & Glb. \\
\midrule
None (reference) & $1.00\times$ & 0.0 & 0.0 & 0.0 \\
Step only & $1.50\times$ & 62.3 & 0.0 & 62.4 \\
Step $+$ coverage (\asCVS) & $2.06\times$ & 51.2 & 51.0 & 39.4 \\
Step $+$ path (\asWAB) & $1.67\times$ & 61.5 & 0.0 & 92.6 \\
\rowcolor{oursbg}\textbf{\method} & \best{$2.59\times$} & 53.2 & 51.5 & 92.6 \\
\bottomrule
\end{tabular}
\end{table}

\noindent\textbf{Matched fidelity.}
A single operating point can favor either side of the speed--fidelity trade-off, so we also compare speedups at equal PSNR (\cref{tab:isoquality}).
\method is ahead across the whole 28--34\,dB band, with an AUC of $2.47$ against $1.93$ for EasyCache ($+28\%$) and $2.27\times$ against $1.77\times$ at 32\,dB.
Path-uniform keeps part of the gain (AUC $2.11$) and step-only keeps little ($1.57$): the advantage comes from allocating reuse according to the structure of the loop rather than from more aggressive step reuse.

\begin{figure*}[t]
  \centering
  \includegraphics[width=\textwidth]{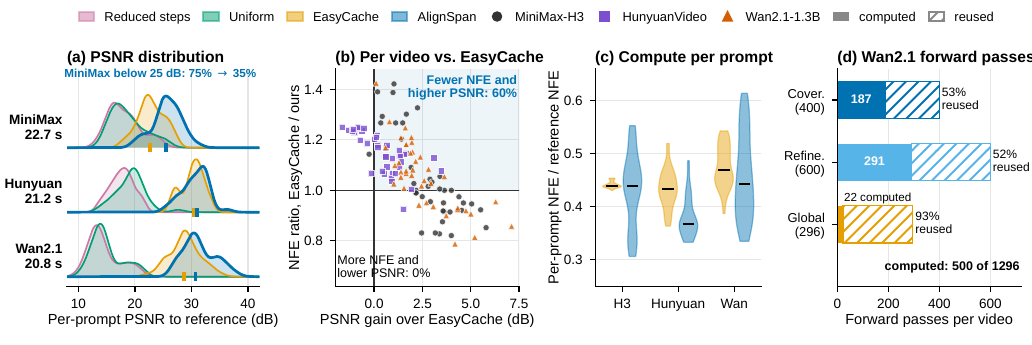}
  \vspace{-14pt}
  \caption{\textbf{Per-video analysis on VBench-Long} (a--c, 40 prompts per backbone, $N{=}4$) and the fate of reference forwards (d).
  (a)~Per-prompt PSNR (ticks: EasyCache and \method medians).
  (b)~PSNR gain and NFE ratio over EasyCache for every video (NFE avoids the timing shift of the MiniMax-H3 EasyCache re-run).
  (c)~Forwards per prompt relative to the reference.
  (d)~Computed and reused forwards of a 20.8\,s Wan2.1 video at 31.1\,dB (\cref{tab:axis}).}
  \label{fig:insights}
  \vspace{-4pt}
\end{figure*}

\noindent\textbf{Contribution of each axis.}
\Cref{tab:axis} decomposes the speedup at a fixed 31.1\,dB.
Step-only reuse removes covering and global passes at the same rate ($62\%$), as expected from a scheduler that treats all units alike.
Adding \asCVS opens up the refinement windows ($51\%$ reused) and raises the speedup to $2.06\times$; adding \asWAB on top raises global reuse from $39\%$ to $93\%$ and the speedup to $2.59\times$ at the same fidelity.
Covering windows, which carry the primary content, are reused at about $53\%$ in both cases.
Of the 1296 forwards of the reference, 500 remain (\cref{fig:insights}d).

\subsection{Analysis}
\label{sec:analysis}

\noindent\textbf{The gain holds per video.}
The PSNR distribution of \method is shifted to the right on all three backbones, most clearly in the lower tail, where caching errors are most visible: the 10th-percentile PSNR rises from $20.0$ to $22.2$\,dB on MiniMax-H3 and from $25.4$ to $27.6$\,dB on Wan2.1, and the worst Wan2.1 video improves from $22.3$ to $27.3$\,dB.
Compared video by video (\cref{fig:insights}b, \cref{tab:pervideo}), \method is more faithful than EasyCache on $70$--$100\%$ of the prompts of each backbone and never uses more forwards while being less faithful.
Within each backbone the PSNR gain is negatively correlated with the NFE ratio ($r={-}0.68$ to ${-}0.77$): the prompts on which \method spends more forwards than EasyCache are those on which it gains the most fidelity.
\Cref{fig:insights}c shows the same behavior from the compute side: on MiniMax-H3, EasyCache uses $43$--$45\%$ of the reference forwards for every prompt, whereas \method uses between $31\%$ and $55\%$, depending on how well each prompt tolerates reuse during the denoising process.

\noindent\textbf{Is it only a better step cache?}
For image generation ($N{=}1$) there are no refinement windows and no separate global path, so \asWAB and \asCVS have nothing to act on and \method reduces to a calibrated, drift-corrected step cache.
In this setting \method is still the most faithful accelerator on HunyuanVideo and Wan2.1, but it is \emph{slower} than EasyCache on both (\cref{tab:short}).
The speed advantage at $N{=}4$ therefore comes from the structure of the long-video loop.

\begin{table}[t]
\centering
\caption{\textbf{Scaling with video length} on Wan2.1 (16\,fps): end-to-end speedup at a fixed 30\,dB ($n{=}10$) for 10.3\,s to 83.8\,s videos. Uniform interval never reaches 30\,dB (16.8--18.1\,dB); $^\dagger$ marks its $q{=}2$ points. Slope is the change per doubling of $N$.}
\label{tab:scaling}
\small
\setlength{\tabcolsep}{3.0pt}
\begin{tabular}{lccccc}
\toprule
\multirow{2}{*}{\textbf{Method}} & $N{=}2$ & $N{=}4$ & $N{=}8$ & $N{=}16$ & \multirow{2}{*}{Slope} \\
& 10.3\,s & 20.8\,s & 41.8\,s & 83.8\,s & \\
\midrule
Uniform ($q{=}2$) & $1.89^\dagger$ & $1.83^\dagger$ & $1.96^\dagger$ & $1.96^\dagger$ & \na \\
Per-unit adaptive & 2.08 & 2.05 & 2.10 & 2.06 & 0.00 \\
Step-only (abl.) & 1.66 & 1.54 & 1.53 & 1.52 & $-0.01$ \\
\rowcolor{oursbg}\textbf{\method}& \best{2.73} & \best{2.76} & \best{2.81} & \best{2.75} & 0.00 \\
\bottomrule
\end{tabular}
\end{table}

\begin{table}[t]
\centering
\caption{\textbf{Ablation} on Wan2.1, $N{=}4$, 20.8\,s videos, shared budget $\theta{=}0.21$ ($n{=}10$). Best speedup and fidelity in \textbf{bold}. Seam: steps with $J_t{>}J_{\max}$ per video. Variants ran on a machine ${\sim}7\%$ slower per forward; the w/o-probe and no-seam-gate rows have the same NFE as the full method.}
\label{tab:ablation}
\small
\setlength{\tabcolsep}{2.25pt}
\begin{tabular}{lcccc}
\toprule
\textbf{Configuration} & \textbf{Speedup} & \textbf{PSNR}$\uparrow$ & \textbf{LPIPS}$\downarrow$ & \textbf{Seam} \\
\midrule
\rowcolor{oursbg}\method (full) & $2.52\times$ & 31.22 & 0.05 & 0.0 \\
\quad w/o \asWAB (path-uniform) & $3.03\times$ & 24.94 & 0.14 & 0.5 \\
\quad w/o \asCVS (windows equal) & $1.49\times$ & \best{33.17} & \best{0.03} & 0.0 \\
\quad w/o probe (open loop) & $2.38\times$ & 31.22 & 0.05 & 0.0 \\
\quad \asWAB, inverted allocation & \best{$3.94\times$} & 16.37 & 0.52 & 0.0 \\
\quad \asCVS without seam gate & $2.39\times$ & 31.22 & 0.05 & 0.0 \\
\bottomrule
\end{tabular}
\end{table}

\begin{table}[t]
\centering
\caption{\textbf{Per-video comparison with EasyCache} on VBench-Long (40 prompts per backbone): share of prompts where \method uses fewer forwards (NFE), has higher PSNR, lower LPIPS, or both fewer forwards and higher PSNR; median PSNR gain ($\Delta$dB). $^\ddagger$Remaining prompts tie in NFE. Full version: supp.\ \cref{tab:pervideovbl}.}
\label{tab:pervideo}
\small
\setlength{\tabcolsep}{3.0pt}
\begin{tabular}{lccccc}
\toprule
\textbf{Backbone} & \textbf{NFE}$\downarrow$ & \textbf{PSNR}$\uparrow$ & \textbf{LPIPS}$\downarrow$ & \textbf{Both} & \textbf{$\Delta$dB} \\
\midrule
MiniMax-H3 & 45\%$^\ddagger$ & 98\% & 92\% & 42\% & $+2.90$ \\
HunyuanVideo & 98\% & 70\% & 70\% & 68\% & $+0.58$ \\
Wan2.1, $\theta{=}0.21$ & 70\% & 100\% & 100\% & 70\% & $+2.00$ \\
Wan2.1, $\theta{=}0.24$ & 95\% & 85\% & 88\% & 80\% & $+0.94$ \\
\bottomrule
\end{tabular}
\end{table}

\begin{table}[t]
\centering
\caption{\textbf{Image generation} ($N{=}1$, 100 prompts): speedup and PSNR (dB) to the unaccelerated output; best in \textbf{bold}. MiniMax-H3 uses 9 frames. Full version: supp.\ \cref{tab:shortfull}.}
\label{tab:short}
\small
\setlength{\tabcolsep}{3.4pt}
\begin{tabular}{lcccccc}
\toprule
& \multicolumn{2}{c}{\textbf{MiniMax-H3}} & \multicolumn{2}{c}{\textbf{Hunyuan}} & \multicolumn{2}{c}{\textbf{Wan2.1}} \\
\cmidrule(lr){2-3}\cmidrule(lr){4-5}\cmidrule(lr){6-7}
\textbf{Method} & Speed & PSNR & Speed & PSNR & Speed & PSNR \\
\midrule
Reduced & $4.88\times$ & 21.65 & \best{$1.98\times$} & 21.93 & \best{$2.19\times$} & 16.93 \\
Uniform & $3.99\times$ & \best{22.52} & $1.73\times$ & 23.63 & $1.83\times$ & 16.76 \\
EasyCache & $4.42\times$ & 20.65 & $1.88\times$ & 28.78 & $2.17\times$ & 22.95 \\
\rowcolor{oursbg}\textbf{\method} & \best{$4.92\times$} & 22.22 & $1.75\times$ & \best{32.76} & $1.79\times$ & \best{29.24} \\
\bottomrule
\end{tabular}
\end{table}

\noindent\textbf{Length and ablation.}
The matched-30\,dB speedup of \method stays at $2.73$--$2.81\times$ from $N{=}2$ to $16$ (\cref{tab:scaling}), so the structural savings do not shrink as videos get longer.
In \cref{tab:ablation}, removing \asWAB or inverting its allocation buys speed at a large loss of fidelity ($24.9$ and $16.4$\,dB), so the direction of the split matters, and removing \asCVS restores fidelity but drops the speedup to $1.49\times$.
Removing the probe or the seam gate does not change any reuse decision at this budget: both variants reproduce the NFE, PSNR, and LPIPS of the full method, indicating that the measured error stayed below the budget and no seam exceeded $J_{\max}$.
They act as safeguards for more aggressive budgets.

\section{Conclusion and Outlook}
\label{sec:conclusion}

Dual-path long-video sampling costs $(m{+}1)N{-}(m{-}1)$ clip evaluations per step, $14\times$ to $62\times$ a short clip for $N{=}4$ to $16$ on Wan2.1, and these evaluations differ in how much reuse error they tolerate.
The fusion weight determines how strongly the error of each path reaches the output, and only the covering windows carry primary content, while refinement windows smooth the boundaries between them.
\method uses both properties in a calibrated cache scheduler that, on VBench-Long, is faster and more faithful than per-unit adaptive caching on three backbones, and keeps its matched-fidelity speedup as videos grow from 10 to 84\,s; a 21\,s HunyuanVideo sample drops from 2.9\,h to 1.1\,h.
More broadly, an accelerator for a composite sampler should account for how each backbone evaluation enters the final output, and measuring fidelity against the unaccelerated output makes such comparisons precise.

Several directions follow from this work.
The seam test currently operates on latent statistics; a decoder-aware or perceptual seam measure could let refinement reuse be pushed further without visible transitions, which matters most for long videos with many boundaries.
Weight-aware budgeting applies to any convex combination of denoising paths, such as region- or prompt-wise guidance, while covering-set skipping carries over to other samplers that blend overlapping windows, including streaming and autoregressive extension.
The budget $\theta$ is currently set per backbone; choosing it automatically for a target fidelity would remove this last manual step.
Cheaper long-video sampling also makes generated videos more practical as synthetic training data, a direction explored for text through tree-guided data synthesis~\cite{wang2026treesynth}.
Beyond video, generative and language models are increasingly used for scientific design~\cite{jiang2025artificial,wang2025large,jiang2025biological}, including functional RNA design~\cite{huang2026programmable,jiang2026rmsagen,jiang2026utilizing,jiang2025rbptool}, protein design~\cite{li2026ds,zhou2025lm2protein}, and RNA foundation models scaled through iterative refinement~\cite{chen2026scaling}; whether calibrated, structure-aware reuse transfers to such samplers is an open question.
Finally, caching is orthogonal to sparse attention and fewer-step solvers, and combining them is a natural next step toward interactive generation of minute-long videos.



{
    \small
    \bibliographystyle{ieeenat_fullname}
    \bibliography{main}
}

\clearpage
\maketitlesupplementary
\appendix
\renewcommand{\thesection}{\Alph{section}}
\setcounter{table}{0}
\setcounter{figure}{0}
\setcounter{equation}{0}
\renewcommand{\thetable}{S\arabic{table}}
\renewcommand{\thefigure}{S\arabic{figure}}
\renewcommand{\theequation}{S\arabic{equation}}
\renewcommand{\theHtable}{S\arabic{table}}
\renewcommand{\theHfigure}{S\arabic{figure}}
\renewcommand{\theHequation}{S\arabic{equation}}
\noindent This appendix supplements the main paper. Tables, figures, and equations of the appendix are numbered S1, S2, \etc.
It contains method and experimental details (\cref{app:impl,app:exp}), further results on the VBench-Long panel (\cref{app:vbl}), the full results of the 100-prompt VBench panel (\cref{app:vb100}), the full ablation and the image-generation experiment (\cref{app:abl}), and qualitative comparisons (\cref{app:qual}).

\section{Method Details}
\label{app:impl}

\noindent\textbf{Cache indexing.}
The cache is indexed by (kind, unit) and holds, per unit, the latest velocity and the input clip used for drift correction (\cref{eq:drift}). A request prices the unit's own entry with \cref{eq:transfer} and serves it if the estimate is within the threshold. Entries are overwritten whenever the unit is computed exactly.

\noindent\textbf{Threshold caps.}
The thresholds of \cref{eq:wab} diverge as $c_t\to0$ or $c_t\to1$, so both are capped at $\theta_{\max}$. Besides numerical safety, the cap prevents a path from being served from cache for the rest of the trajectory because its weight is small, which would leave its structure unenforced at steps where the weight is small but not negligible.

\noindent\textbf{Seam gate.}
The seam neighbourhood $\mathcal{S}$ uses a fixed radius around each covering boundary, and the interior set $\mathcal{I}$ excludes these neighbourhoods. $J_t$ (\cref{eq:seam}) is averaged over channels and computed on the updated latent $\bm{Z}_{t-1}$, so the decision at the next step uses the most recent sequence.

\noindent\textbf{Probe eligibility.}
Budget updates are suppressed at the earliest steps, where the latent is dominated by noise and the probe is uninformative, and at the final steps. The error tables $e_{\mathrm{step}}$ are updated on every exact computation regardless.

\noindent\textbf{Default hyperparameters (Wan2.1 and HunyuanVideo).}
Budget $\theta{=}0.21$, threshold cap $\theta_{\max}{=}2.0$, $J_{\max}{=}1.25$, $j_{\mathrm{margin}}{=}1.05$, safety margin $\eta{=}0.005$, $\lambda_{\mathrm{std}}{=}2$, $n_{\min}{=}1$, feedback gain $\beta{=}0.2$, $c_{\mathrm{clip}}{=}0.5$, $s_t\in[0.25,1]$, eligibility window $[0.1,0.9]$, anchor stride $s_{\mathrm{anc}}{=}3$, age extrapolation and drift-corrected reuse enabled. MiniMax-H3 uses $\theta{=}0.26$, $\lambda_{\mathrm{std}}{=}0.5$, $s_{\mathrm{anc}}{=}7$, and $\theta_{\max}{=}0.5$. The additional Wan2.1 row of \cref{tab:main} changes only the budget, to $\theta{=}0.24$.

\section{Experimental Details}
\label{app:exp}

\noindent\textbf{Configurations.}
The window length equals the native latent length of each backbone. Per-backbone $(U,s,m,N)$, budget, anchor stride, EasyCache settings, and feedback gains are the same for all experiments that use a given backbone, except for the budget sweeps. Hyperparameters were tuned on prompts that overlap the 100-prompt panel and the 10-prompt suite; there is no separate held-out set for these two.

\noindent\textbf{VBench-Long panel.}
The panel has 40 long, descriptive prompts (48 words on average; \eg, \emph{``A stylish woman walks down a Tokyo street filled with warm glowing neon and animated city signage\ldots''}). Each video is evaluated with the VBench-Long protocol, which splits it into short clips (12 per video on MiniMax-H3, 11--13 on HunyuanVideo, 11 on Wan2.1) and scores each clip. For each video we average the clip scores per dimension and form three groups:
\begin{equation}
\begin{aligned}
\text{Temporal}&=\tfrac{1}{3.5}\bigl(\text{Subj.}+\text{Bkg.}+\text{Motion}+0.5\,\text{Dyn.}\bigr),\\
\text{Frame}&=\tfrac12\bigl(\text{Aesthetic}+\text{Imaging}\bigr),\\
\text{Text}&=\tfrac12\bigl(\text{Overall consistency}+\text{CLIP}\bigr).
\end{aligned}
\label{eq:vblgroups}
\end{equation}
Panel scores are clip-weighted means over videos, which is how VBench-Long pools clips.

\noindent\textbf{Fidelity measurement.}
All fidelity metrics compare against the unaccelerated dual-path output with the same seed, prompt, and pipeline configuration, decoded to pixels before computing PSNR, SSIM, and LPIPS, the standard full-reference measures also used in image reconstruction~\cite{jiang2026thz}. On the 100-prompt panel, the MiniMax-H3 uniform-interval fidelity mixes two host schedules for the global path across the two halves of the prompts; the printed numbers are those of the mixed reference campaign.

\noindent\textbf{Timing.}
Latency is end-to-end wall-clock time including decoding (and prompt encoding on HunyuanVideo and Wan2.1). Speedups are ratios of mean latencies. On the VBench-Long panel, overhead is computed from denoising time as in \cref{tab:main}, with reference times per forward of $8.77$, $22.21$, and $1.59$\,s (MiniMax-H3, HunyuanVideo, Wan2.1). On the 100-prompt panel (\cref{tab:vb100full}) it is computed from end-to-end time, with $9.95$, $24.67$, and $1.72$\,s per forward. The two panels were run in separate campaigns, and absolute times differ between them by up to $11\%$ for the same configuration (\eg, $2.88$ vs.\ $3.20$\,h for a HunyuanVideo reference video), so we compare speedups only within a panel. In the VBench-Long campaign, EasyCache on MiniMax-H3 was re-run in a later session with the same configuration; the reference was not re-run. Normalized by the per-forward time of each session, that row corresponds to $2.22\times$ and $+0.70\%$ overhead instead of $2.08\times$ and $+7.37\%$. The sampler records only whole-step times; the global path accounts for about $29\%$ of the latency of a gate-open step.

\section{Additional Results on VBench-Long}
\label{app:vbl}

\begin{table*}[t]
\centering
\caption{\textbf{VBench-Long dimensions} on the 40-prompt panel (clip-weighted means; $N{=}4$). These scores are reference-free and not ranked; the reference row is the unaccelerated output. Subj./Bkg.: subject and background consistency; Motion: motion smoothness; Dyn.: dynamic degree; Aesth./Imag.: aesthetic and imaging quality; Overall: overall consistency.}
\label{tab:vbldims}
\small
\setlength{\tabcolsep}{4.2pt}
\begin{tabular}{ll cccc cc cc}
\toprule
& & \multicolumn{4}{c}{\textbf{Temporal}} & \multicolumn{2}{c}{\textbf{Frame}} & \multicolumn{2}{c}{\textbf{Text}} \\
\cmidrule(lr){3-6}\cmidrule(lr){7-8}\cmidrule(lr){9-10}
\textbf{Backbone} & \textbf{Method} & Subj. & Bkg. & Motion & Dyn. & Aesth. & Imag. & Overall & CLIP \\
\midrule
\multirow{5}{*}{\shortstack[l]{MiniMax-H3\\22.7\,s}} & Reference & 0.9562 & 0.9690 & 0.9909 & 0.6521 & 0.5959 & 0.6574 & 0.2700 & 0.3332 \\
& Reduced steps & 0.9596 & 0.9714 & 0.9918 & 0.6083 & 0.6064 & 0.6492 & 0.2711 & 0.3332 \\
& Uniform interval & 0.9578 & 0.9698 & 0.9913 & 0.6104 & 0.6012 & 0.6534 & 0.2708 & 0.3337 \\
& EasyCache & 0.9434 & 0.9627 & 0.9886 & 0.5437 & 0.5583 & 0.6276 & 0.2619 & 0.3279 \\
& \cellcolor{oursbg}\method & \cellcolor{oursbg}0.9511 & \cellcolor{oursbg}0.9664 & \cellcolor{oursbg}0.9897 & \cellcolor{oursbg}0.5854 & \cellcolor{oursbg}0.5844 & \cellcolor{oursbg}0.6453 & \cellcolor{oursbg}0.2673 & \cellcolor{oursbg}0.3309 \\
\midrule
\multirow{5}{*}{\shortstack[l]{HunyuanVideo\\21.2\,s}} & Reference & 0.9599 & 0.9726 & 0.9919 & 0.4676 & 0.6105 & 0.6111 & 0.2735 & 0.3402 \\
& Reduced steps & 0.9564 & 0.9708 & 0.9928 & 0.4547 & 0.5899 & 0.5533 & 0.2711 & 0.3343 \\
& Uniform interval & 0.9562 & 0.9722 & 0.9919 & 0.4254 & 0.6022 & 0.5825 & 0.2718 & 0.3372 \\
& EasyCache & 0.9576 & 0.9702 & 0.9914 & 0.3714 & 0.6019 & 0.6087 & 0.2722 & 0.3394 \\
& \cellcolor{oursbg}\method & \cellcolor{oursbg}0.9584 & \cellcolor{oursbg}0.9715 & \cellcolor{oursbg}0.9918 & \cellcolor{oursbg}0.3893 & \cellcolor{oursbg}0.6056 & \cellcolor{oursbg}0.6073 & \cellcolor{oursbg}0.2727 & \cellcolor{oursbg}0.3400 \\
\midrule
\multirow{6}{*}{\shortstack[l]{Wan2.1-1.3B\\20.8\,s}} & Reference & 0.9565 & 0.9606 & 0.9859 & 0.4795 & 0.5879 & 0.6563 & 0.2616 & 0.3338 \\
& Reduced steps & 0.9495 & 0.9567 & 0.9819 & 0.4347 & 0.5531 & 0.6113 & 0.2394 & 0.3179 \\
& Uniform interval & 0.9510 & 0.9583 & 0.9830 & 0.4086 & 0.5562 & 0.6094 & 0.2424 & 0.3218 \\
& EasyCache & 0.9557 & 0.9606 & 0.9861 & 0.4364 & 0.5845 & 0.6519 & 0.2618 & 0.3335 \\
& \cellcolor{oursbg}\method ($\theta{=}0.21$) & \cellcolor{oursbg}0.9562 & \cellcolor{oursbg}0.9608 & \cellcolor{oursbg}0.9860 & \cellcolor{oursbg}0.4591 & \cellcolor{oursbg}0.5868 & \cellcolor{oursbg}0.6527 & \cellcolor{oursbg}0.2617 & \cellcolor{oursbg}0.3336 \\
& \method ($\theta{=}0.24$) & 0.9567 & 0.9610 & 0.9862 & 0.4477 & 0.5870 & 0.6512 & 0.2616 & 0.3334 \\
\bottomrule
\end{tabular}
\end{table*}

\noindent\textbf{Per-dimension scores.}
\Cref{tab:vbldims} lists the eight VBench-Long dimensions behind the group scores of \cref{tab:main}.
Consistency and motion smoothness change by at most $0.013$ for any accelerator.
The dimension that moves most is dynamic degree, which all accelerators lower relative to the reference; the drop is smaller for \method than for EasyCache on all three backbones ($0.652\to0.585$ vs.\ $0.544$ on MiniMax-H3, $0.468\to0.389$ vs.\ $0.371$ on HunyuanVideo, $0.480\to0.459$ vs.\ $0.436$ on Wan2.1).
On MiniMax-H3, reduced steps and uniform caching keep consistency and frame quality close to the reference and their dynamic degree above that of both caches, although their videos differ strongly from the reference (18--19\,dB); this is why we do not rank methods by these scores. In the same spirit as trajectory-aware evaluation of agents~\cite{chen2026trace}, which looks beyond final-outcome scores, we judge accelerators by their fidelity to the reference trajectory's output rather than by reference-free quality alone.

\begin{table}[t]
\centering
\caption{\textbf{Per-video comparison with EasyCache on VBench-Long} (40 prompts per backbone): share of prompts on which \method uses fewer forwards (NFE), is faster in wall-clock time, has higher PSNR or lower LPIPS, and is both cheaper and higher in PSNR; median NFE ratio (EasyCache/ours) and median PSNR gain. No prompt has more NFE and lower PSNR, and none is slower and lower in PSNR. $^\ddagger$Three prompts have equal NFE.}
\label{tab:pervideovbl}
\footnotesize
\setlength{\tabcolsep}{1.55pt}
\begin{tabular}{lcccccccc}
\toprule
& \multicolumn{2}{c}{\textbf{Cheaper}} & \multicolumn{2}{c}{\textbf{Better}} & \multicolumn{2}{c}{\textbf{Both}} & \multicolumn{2}{c}{\textbf{Median}} \\
\cmidrule(lr){2-3}\cmidrule(lr){4-5}\cmidrule(lr){6-7}\cmidrule(lr){8-9}
\textbf{Backbone} & NFE & Time & PSNR & LPIPS & NFE & Time & NFE$\times$ & $\Delta$dB \\
\midrule
MiniMax-H3 & 45\%$^\ddagger$ & 70\% & 98\% & 92\% & 42\% & 68\% & 1.00 & $+2.90$ \\
HunyuanVideo & 98\% & 95\% & 70\% & 70\% & 68\% & 65\% & 1.15 & $+0.58$ \\
Wan2.1 ($\theta{=}0.21$) & 70\% & 70\% & 100\% & 100\% & 70\% & 70\% & 1.06 & $+2.00$ \\
Wan2.1 ($\theta{=}0.24$) & 95\% & 95\% & 85\% & 88\% & 80\% & 80\% & 1.19 & $+0.94$ \\
\midrule
All 120 ($\theta{=}0.21$) & 71\% & 78\% & 89\% & 88\% & 60\% & 68\% & \na & \na \\
\bottomrule
\end{tabular}
\end{table}

\noindent\textbf{Per-video comparison.}
\Cref{tab:pervideovbl} summarizes the video-by-video comparison behind \cref{fig:insights}b.
On MiniMax-H3 the two methods use almost the same number of forwards (median ratio $1.00$), and the advantage of \method is in fidelity: higher PSNR on 39 of 40 prompts with a median gain of $2.9$\,dB.
On HunyuanVideo it uses fewer forwards on 39 of 40 prompts, and on Wan2.1 it has higher PSNR and lower LPIPS on every prompt.
With the looser budget $\theta{=}0.24$, \method on Wan2.1 is faster than EasyCache on 38 of 40 prompts and still more faithful on 34.

\begin{figure*}[t]
  \centering
  \includegraphics[width=\textwidth]{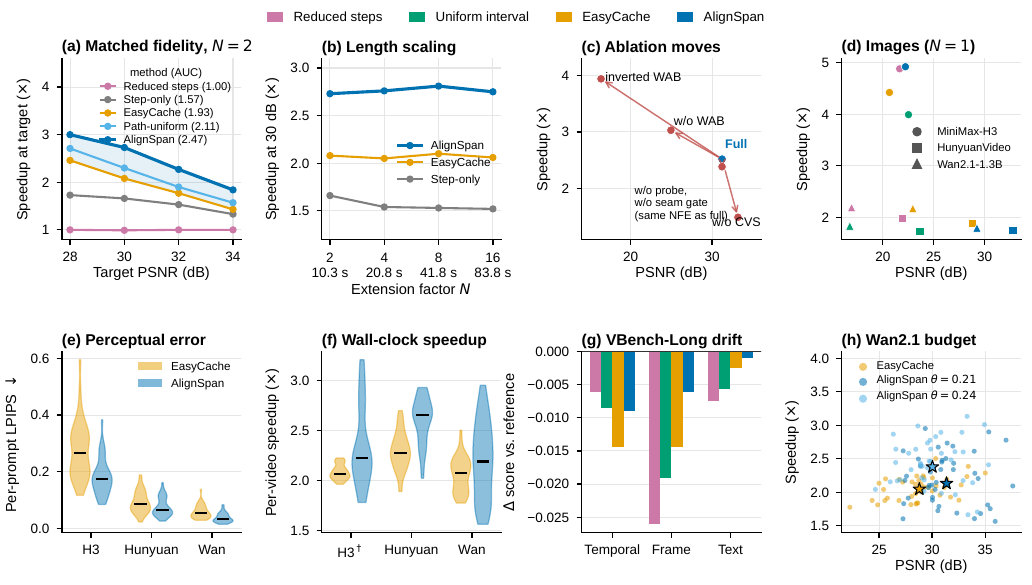}
  \vspace{-12pt}
  \caption{\textbf{Additional analyses.} Top row, from the 10-prompt Wan2.1 suite and the image-generation runs:
  (a)~matched-fidelity curves of \cref{tab:isoquality} (10.3\,s videos), with the gain of \method over EasyCache shaded;
  (b)~speedup at 30\,dB versus extension factor (\cref{tab:scaling});
  (c)~ablation variants of \cref{tab:ablationfull} as moves away from the full method (20.8\,s videos);
  (d)~image generation (\cref{tab:shortfull}; colour is the method, marker shape the backbone).
  Bottom row, from the 40-prompt VBench-Long panel:
  (e)~per-prompt LPIPS (bars: medians);
  (f)~per-video wall-clock speedup ($^\dagger$EasyCache on MiniMax-H3 was re-run in a slower session, which lowers its speedups by about $7\%$);
  (g)~change of the three VBench-Long group scores relative to the reference, averaged over backbones;
  (h)~Wan2.1 per-video speedup and PSNR for EasyCache and \method at two budgets (stars: panel means).}
  \label{fig:more}
\end{figure*}

\noindent\textbf{Distributions and budgets.}
\Cref{fig:more}e--h complements \cref{fig:insights}.
Per-prompt LPIPS improves at the median ($0.265\to0.174$, $0.086\to0.064$, $0.054\to0.033$ on MiniMax-H3, HunyuanVideo, Wan2.1) and at the 90th percentile ($0.379\to0.269$, $0.142\to0.119$, $0.089\to0.058$).
Per-video speedups have a higher median than EasyCache on every backbone ($2.22$ vs.\ $2.06$, $2.66$ vs.\ $2.27$, $2.19$ vs.\ $2.07$) and a wider spread, because the per-sample error model reuses more aggressively on prompts that tolerate it.
On Wan2.1 (\cref{fig:more}h), raising the budget from $0.21$ to $0.24$ moves the mean operating point from $2.13\times$ at $31.4$\,dB to $2.38\times$ at $30.0$\,dB; both points lie above and to the right of EasyCache ($2.05\times$ at $28.8$\,dB).

\noindent\textbf{Frontiers and ablation moves.}
\Cref{fig:more}a shows that the gain over EasyCache holds across the whole 28--34\,dB band and that path-uniform budgeting recovers only part of it.
In \cref{fig:more}c, every component removal that increases speed (w/o \asWAB, inverted \asWAB) does so by losing fidelity, while removing \asCVS moves in the opposite direction.

\begin{table}[t]
\centering
\caption{\textbf{Generation time} on one RTX~5090 ($N{=}4$): hours per video (mean), hours per minute of generated video, and GPU-hours for the whole panel. The two panels were run in separate campaigns, so absolute times are comparable only within a panel.}
\label{tab:time}
\footnotesize
\setlength{\tabcolsep}{2.6pt}
\begin{adjustbox}{max width=\linewidth}
\begin{tabular}{llcccccc}
\toprule
& & \multicolumn{3}{c}{\textbf{VBench-Long (40)}} & \multicolumn{3}{c}{\textbf{VBench (100)}} \\
\cmidrule(lr){3-5}\cmidrule(lr){6-8}
& \textbf{Method} & h/video & h/min & GPU-h & h/video & h/min & GPU-h \\
\midrule
\multirow{3}{*}{\shortstack{H3\\22.7\,s}} & Reference & 1.07 & 2.83 & 42.7 & 1.21 & 3.21 & 121.1 \\
& EasyCache & 0.51$^\dagger$ & 1.36$^\dagger$ & 20.5$^\dagger$ & 0.54 & 1.42 & 53.5 \\
& \cellcolor{oursbg}\method & \cellcolor{oursbg}\textbf{0.47} & \cellcolor{oursbg}\textbf{1.24} & \cellcolor{oursbg}\textbf{18.8} & \cellcolor{oursbg}\textbf{0.48} & \cellcolor{oursbg}\textbf{1.27} & \cellcolor{oursbg}\textbf{47.9} \\
\midrule
\multirow{3}{*}{\shortstack{Hy.\\21.2\,s}} & Reference & 2.88 & 8.16 & 115.3 & 3.20 & 9.05 & 320.0 \\
& EasyCache & 1.27 & 3.58 & 50.6 & 1.35 & 3.82 & 135.1 \\
& \cellcolor{oursbg}\method & \cellcolor{oursbg}\textbf{1.12} & \cellcolor{oursbg}\textbf{3.16} & \cellcolor{oursbg}\textbf{44.7} & \cellcolor{oursbg}\textbf{1.19} & \cellcolor{oursbg}\textbf{3.36} & \cellcolor{oursbg}\textbf{118.9} \\
\midrule
\multirow{4}{*}{\shortstack{Wan\\20.8\,s}} & Reference & 0.58 & 1.67 & 23.1 & 0.62 & 1.80 & 62.4 \\
& EasyCache & 0.28 & 0.82 & 11.3 & 0.29 & 0.83 & 28.9 \\
& \cellcolor{oursbg}\method & \cellcolor{oursbg}0.27 & \cellcolor{oursbg}0.78 & \cellcolor{oursbg}10.9 & \cellcolor{oursbg}\textbf{0.27} & \cellcolor{oursbg}\textbf{0.79} & \cellcolor{oursbg}\textbf{27.4} \\
& \method, $\theta{=}0.24$ & \textbf{0.24} & \textbf{0.70} & \textbf{9.7} & \na & \na & \na \\
\bottomrule
\end{tabular}
\end{adjustbox}
\vspace{2pt}

{\footnotesize $^\dagger$Re-run in a slower session (\cref{app:exp}).}
\end{table}

\noindent\textbf{Generation time.}
\Cref{tab:time} converts the latencies into the units a practitioner budgets in.
On VBench-Long, one minute of generated video costs $2.83$, $8.16$, and $1.67$ GPU-hours on an RTX~5090 without acceleration (MiniMax-H3, HunyuanVideo, Wan2.1), and $1.24$, $3.16$, and $0.78$ hours with \method.

\FloatBarrier
\section{Full Results on the 100-Prompt VBench Panel}
\label{app:vb100}

\begin{table*}[t]
\centering
\caption{\textbf{Full results on the 100-prompt VBench panel} ($N{=}4$, seed 0, one RTX~5090 per video; compact version in \cref{tab:vbench100}). Fidelity is measured against the unaccelerated dual-path output. Best and second-best accelerators in efficiency and fidelity are in \best{bold} and \second{underline}; VBench and resource columns are not ranked. Overhead is $(t_{\mathrm{e2e}}-\mathrm{NFE}\cdot t_{\mathrm{fwd}})/(\mathrm{NFE}\cdot t_{\mathrm{fwd}})$ and is not comparable with the denoising-time overhead of \cref{tab:main}.}
\label{tab:vb100full}
\small
\setlength{\tabcolsep}{3.35pt}
\begin{adjustbox}{max width=\textwidth}
\begin{tabular}{l cc ccc ccc cc}
\toprule
\multirow{2}{*}{\textbf{Method}} & \multicolumn{2}{c}{\textbf{Efficiency}} & \multicolumn{3}{c}{\textbf{Fidelity to reference}} & \multicolumn{3}{c}{\textbf{VBench (not ranked)}} & \multicolumn{2}{c}{\textbf{Resource (not ranked)}} \\
\cmidrule(lr){2-3}\cmidrule(lr){4-6}\cmidrule(lr){7-9}\cmidrule(lr){10-11}
& Lat.\,(s)$\downarrow$ & Speedup$\uparrow$ & PSNR$\uparrow$ & SSIM$\uparrow$ & LPIPS$\downarrow$ & Quality & Semantic & Total & Peak (GiB) & Overhead (\%) \\
\midrule
\multicolumn{11}{l}{\emph{MiniMax-H3: 22.7\,s videos (544 frames, 24\,fps, $960{\times}544$)}}\\
Unaccelerated reference & 4361.1 & $1.00\times$ & \na & \na & \na & 0.8419 & 0.7434 & 0.8222 & 16.26 & \na \\
Reduced steps ($T'{=}22$) & 1939.0 & $2.25\times$ & 18.78 & 0.705 & 0.310 & 0.8380 & 0.7362 & 0.8176 & 16.27 & $+0.32$ \\
Uniform interval ($q{=}2$) & 2363.3 & $1.85\times$ & 19.41 & 0.719 & 0.285 & 0.8393 & 0.7430 & 0.8200 & 16.35 & $+1.39$ \\
Per-unit adaptive (EasyCache) & \second{1926.9} & \second{$2.26\times$} & \second{24.65} & \second{0.772} & \second{0.234} & 0.8081 & 0.7252 & 0.7915 & 16.56 & $-0.34$ \\
\rowcolor{oursbg}\textbf{\method (ours)} & \best{1722.7} & \best{$2.53\times$} & \best{26.70} & \best{0.814} & \best{0.181} & 0.8172 & 0.7350 & 0.8007 & 16.40 & $+0.03$ \\
\midrule
\multicolumn{11}{l}{\emph{HunyuanVideo: 21.2\,s videos (509 frames, 24\,fps, $960{\times}544$)}}\\
Unaccelerated reference & 11518.7 & $1.00\times$ & \na & \na & \na & 0.7968 & 0.7008 & 0.7776 & 23.69 & \na \\
Reduced steps ($T'{=}22$) & 5107.9 & $2.26\times$ & 20.48 & 0.783 & 0.313 & 0.7798 & 0.6931 & 0.7624 & 23.74 & $+0.06$ \\
Uniform interval ($q{=}2$) & 6101.9 & $1.89\times$ & 22.56 & 0.823 & 0.240 & 0.7860 & 0.6824 & 0.7653 & 23.88 & $+0.13$ \\
Per-unit adaptive (EasyCache) & \second{4863.8} & \second{$2.37\times$} & \second{31.78} & \second{0.905} & \second{0.111} & 0.7834 & 0.6807 & 0.7629 & 24.42 & $+0.17$ \\
\rowcolor{oursbg}\textbf{\method (ours)} & \best{4279.2} & \best{$2.69\times$} & \best{32.57} & \best{0.925} & \best{0.078} & 0.7886 & 0.6829 & 0.7674 & 23.94 & $+0.49$ \\
\midrule
\multicolumn{11}{l}{\emph{Wan2.1-1.3B: 20.8\,s videos (333 frames, 16\,fps, $832{\times}480$)}}\\
Unaccelerated reference & 2248.0 & $1.00\times$ & \na & \na & \na & 0.8028 & 0.6315 & 0.7685 & 8.03 & \na \\
Reduced steps ($T'{=}22$) & \second{995.6} & \second{$2.26\times$} & 16.30 & 0.612 & 0.396 & 0.7884 & 0.5784 & 0.7464 & 8.03 & $-0.21$ \\
Uniform interval ($q{=}2$) & 1192.3 & $1.89\times$ & 16.84 & 0.631 & 0.376 & 0.7878 & 0.6114 & 0.7525 & 8.24 & $+0.19$ \\
Per-unit adaptive (EasyCache) & 1041.8 & $2.16\times$ & \second{29.62} & \second{0.886} & \second{0.059} & 0.7969 & 0.6249 & 0.7625 & 8.46 & $+0.00$ \\
\rowcolor{oursbg}\textbf{\method (ours)} & \best{986.1} & \best{$2.28\times$} & \best{32.01} & \best{0.916} & \best{0.039} & 0.7998 & 0.6306 & 0.7660 & 8.19 & $+0.49$ \\
\bottomrule
\end{tabular}
\end{adjustbox}
\end{table*}

\begin{table}[t]
\centering
\caption{\textbf{Per-video comparison with EasyCache} on the 100-prompt VBench panel (\cref{tab:vbench100}, 20.8--22.7\,s videos): share of prompts where \method is faster, has higher PSNR, lower LPIPS, or is both faster and higher in PSNR; median speed ratio and PSNR gain.}
\label{tab:pervideo100}
\footnotesize
\setlength{\tabcolsep}{2.0pt}
\begin{tabular}{lcccccc}
\toprule
\textbf{Backbone} & \textbf{Faster} & \textbf{PSNR}$\uparrow$ & \textbf{LPIPS}$\downarrow$ & \textbf{Both} & \textbf{Med.\,$\times$} & \textbf{Med.\,$\Delta$dB} \\
\midrule
H3 (22.7\,s) & 69\% & 93\% & 88\% & \textbf{62\%} & 1.10 & $+1.87$ \\
Hunyuan (21.2\,s) & 98\% & 69\% & 77\% & \textbf{67\%} & 1.15 & $+0.56$ \\
Wan2.1 (20.8\,s) & 65\% & 100\% & 100\% & \textbf{65\%} & 1.07 & $+2.34$ \\
\bottomrule
\end{tabular}
\end{table}

\begin{figure*}[t]
  \centering
  \includegraphics[width=\textwidth]{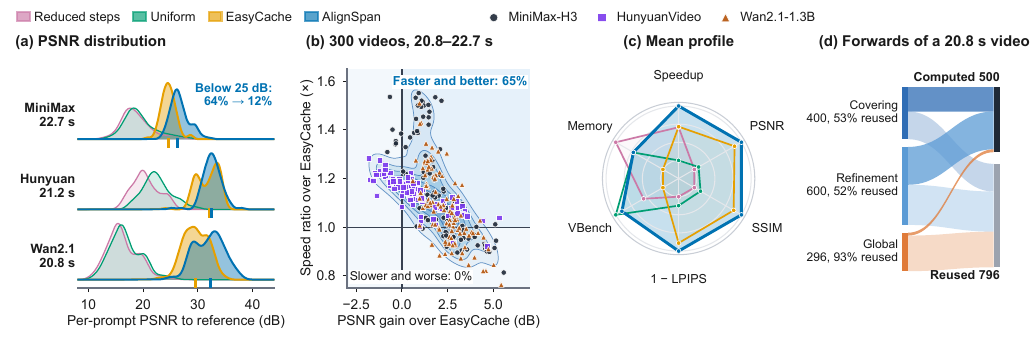}
  \vspace{-6pt}
  \caption{\textbf{Per-video analysis on the 100-prompt VBench panel} (300 videos, 20.8--22.7\,s).
  (a)~Per-prompt PSNR (ticks: medians); on MiniMax-H3 the share of videos below 25\,dB falls from $64\%$ to $12\%$.
  (b)~PSNR gain and wall-clock speed ratio over EasyCache, with density contours; $65\%$ of the videos are both faster and more faithful, and none is both slower and less faithful.
  (c)~Mean profile over backbones (\cref{tab:vb100full}), min--max normalized across accelerators.
  (d)~Fate of the 1296 reference forwards per 20.8\,s Wan2.1 video at a matched 31.1\,dB (\cref{tab:axis}).}
  \label{fig:vb100}
\end{figure*}

\noindent\textbf{Full table and per-video comparison.}
\Cref{tab:vb100full} is the full version of \cref{tab:vbench100}. On this panel \method has the highest speedup and the highest fidelity on all three backbones, and the VBench total of every accelerator stays within $0.031$ of the reference.
\Cref{tab:pervideo100,fig:vb100} give the per-video comparison: the 10th-percentile PSNR rises from $23.2$ to $24.9$\,dB on MiniMax-H3 and from $29.2$ to $30.8$\,dB on HunyuanVideo, and \method is both faster and more faithful than EasyCache on $65\%$ of the 300 videos, with no video that is both slower and less faithful.
Within each backbone the PSNR gain and the speed ratio are negatively correlated ($-0.71$ to $-0.82$), as on VBench-Long.

\FloatBarrier
\section{Ablation and Image Generation}
\label{app:abl}

\begin{table*}[t]
\centering
\caption{\textbf{Full ablation} on Wan2.1 at $N{=}4$, 20.8\,s videos ($n{=}10$), absolute values at a shared budget; best speedup and fidelity in \textbf{bold}. Feedback active is the fraction of steps on which the budget update of \cref{eq:controller} was evaluated. The variants were generated on a machine about $7\%$ slower per forward than the full method; the open-loop and no-seam-gate rows have exactly the same NFE, PSNR, and LPIPS as the full method. Overhead here uses denoising time and is not comparable with \cref{tab:vb100full}.}
\label{tab:ablationfull}
\small
\setlength{\tabcolsep}{4.5pt}
\begin{tabular}{lcccccc}
\toprule
\textbf{Configuration} & \textbf{Speedup}$\uparrow$ & \textbf{PSNR}$\uparrow$ & \textbf{LPIPS}$\downarrow$ & \textbf{Seam viol.} & \textbf{Feedback active} & \textbf{Overhead} \\
\midrule
\rowcolor{oursbg}\method (full, absolute) & $2.52\times$ & 31.22 & 0.05 & 0.0 & 0.42 & 1.43\% \\
\multicolumn{7}{l}{\emph{Component removal}}\\
\quad w/o \asWAB (path-uniform threshold) & $3.03\times$ & 24.94 & 0.14 & 0.5 & 0.40 & 6.70\% \\
\quad w/o \asCVS (all windows equal) & $1.49\times$ & \best{33.17} & \best{0.03} & 0.0 & 0.42 & 7.24\% \\
\quad w/o anchor probe (open-loop thresholds) & $2.38\times$ & 31.22 & 0.05 & 0.0 & 0.00 & 7.15\% \\
\multicolumn{7}{l}{\emph{Design choices within components}}\\
\quad \asWAB with inverted allocation & \best{$3.94\times$} & 16.37 & 0.52 & 0.0 & 0.34 & 6.00\% \\
\quad \asCVS without seam gate ($J_{\max}{=}\infty$) & $2.39\times$ & 31.22 & 0.05 & 0.0 & 0.42 & 7.12\% \\
\bottomrule
\end{tabular}
\end{table*}

\begin{table*}[t]
\centering
\caption{\textbf{Image generation} (single-frame outputs, $N{=}1$, 100 prompts): speedup and fidelity against the unaccelerated image of each backbone. Best accelerator per column in bold.}
\label{tab:shortfull}
\small
\setlength{\tabcolsep}{4pt}
\begin{tabular}{l cccc cccc cccc}
\toprule
& \multicolumn{4}{c}{\textbf{MiniMax-H3} (9 frames, 0.4\,s)} & \multicolumn{4}{c}{\textbf{HunyuanVideo} (1 frame)} & \multicolumn{4}{c}{\textbf{Wan2.1-1.3B} (1 frame)} \\
\cmidrule(lr){2-5}\cmidrule(lr){6-9}\cmidrule(lr){10-13}
\textbf{Method} & Speedup & PSNR & SSIM & LPIPS & Speedup & PSNR & SSIM & LPIPS & Speedup & PSNR & SSIM & LPIPS \\
\midrule
Reduced steps & $4.88\times$ & 21.65 & 0.766 & 0.201 & \best{$1.98\times$} & 21.93 & 0.820 & 0.194 & \best{$2.19\times$} & 16.93 & 0.708 & 0.267 \\
Uniform interval & $3.99\times$ & \best{22.52} & \best{0.797} & \best{0.172} & $1.73\times$ & 23.63 & 0.861 & 0.144 & $1.83\times$ & 16.76 & 0.698 & 0.277 \\
EasyCache & $4.42\times$ & 20.65 & 0.691 & 0.288 & $1.88\times$ & 28.78 & 0.892 & 0.098 & $2.17\times$ & 22.95 & 0.821 & 0.106 \\
\rowcolor{oursbg}\method & \best{$4.92\times$} & 22.22 & 0.752 & 0.217 & $1.75\times$ & \best{32.76} & \best{0.940} & \best{0.042} & $1.79\times$ & \best{29.24} & \best{0.901} & \best{0.053} \\
\bottomrule
\end{tabular}
\end{table*}

\noindent\textbf{Full ablation.}
\Cref{tab:ablationfull} reports all ablation rows with feedback activity and overhead.
The full, open-loop, and no-seam-gate rows have identical NFE, PSNR, and LPIPS, so at this budget neither the budget feedback nor the seam veto changes a reuse decision; their wall-clock speedups differ only through the slower machine used for the variants, and NFE is the hardware-independent comparison for these rows.

\noindent\textbf{Image generation ($N{=}1$).}
To separate the effect of the dual-path structure from that of a better step cache, we ran all methods on image generation with the same backbones (single-frame outputs; MiniMax-H3 uses its shortest supported output of 9 frames; 100 prompts, seed 0, fidelity against the unaccelerated output of the same backbone).
There are no refinement windows and no separate global path, so \asWAB and \asCVS have nothing to act on and \method reduces to a calibrated, drift-corrected step cache.
\method remains the most faithful accelerator on HunyuanVideo ($32.76$ vs.\ $28.78$\,dB) and Wan2.1 ($29.24$ vs.\ $22.95$\,dB) but is slower than EasyCache there ($1.75\times$ vs.\ $1.88\times$ and $1.79\times$ vs.\ $2.17\times$); on MiniMax-H3 it is the fastest method, at a PSNR between those of EasyCache and uniform caching (\cref{tab:shortfull}).
The speed advantage at $N{=}4$ therefore comes from the structure of the long-video loop rather than from the step cache alone.

\FloatBarrier
\section{Qualitative Comparisons}
\label{app:qual}

\begin{table*}[t]
\centering
\caption{\textbf{Per-video speed and fidelity} for the 30 qualitative cases of the 100-prompt VBench panel ($N{=}4$; MiniMax-H3 22.7\,s, HunyuanVideo 21.2\,s, Wan2.1 20.8\,s videos). Each cell is speedup over the reference time of the same prompt / PSNR to the reference (dB). The best speedup and the best PSNR among the four accelerators are in \textbf{bold} for every video.}
\label{tab:cases}
\footnotesize
\setlength{\tabcolsep}{5pt}
\begin{tabular}{llp{3.6cm}rcccc}
\toprule
\textbf{Backbone} & \textbf{ID} & \textbf{Prompt} & \textbf{Ref.\ time} & \textbf{Reduced steps} & \textbf{Uniform} & \textbf{EasyCache} & \textbf{\method} \\
\midrule
H3 (22.7\,s) & 93 & time lapse of sunrise on mars & 4249 (1.18\,h) & 2.13/27.7 & 1.88/29.0 & 2.26/28.8 & \textbf{3.44}/\textbf{31.0} \\
H3 (22.7\,s) & 4 & a person & 4224 (1.17\,h) & 2.19/16.9 & 1.80/17.0 & 2.27/28.3 & \textbf{3.23}/\textbf{29.8} \\
H3 (22.7\,s) & 58 & A person is clapping & 4367 (1.21\,h) & 2.21/19.9 & 1.85/20.4 & 2.27/26.6 & \textbf{2.48}/\textbf{29.9} \\
H3 (22.7\,s) & 90 & a parking meter on the right\ldots & 4365 (1.21\,h) & 2.24/17.6 & 1.81/17.6 & 2.23/24.3 & \textbf{2.26}/\textbf{29.4} \\
H3 (22.7\,s) & 53 & an airplane & 4267 (1.19\,h) & 2.16/17.9 & 1.81/18.7 & \textbf{2.19}/25.5 & \textbf{2.19}/\textbf{29.6} \\
H3 (22.7\,s) & 28 & Sunset time lapse at the bea\ldots & 4450 (1.24\,h) & 2.31/24.7 & 1.84/25.1 & 2.43/26.1 & \textbf{3.64}/\textbf{27.3} \\
H3 (22.7\,s) & 63 & basement & 4355 (1.21\,h) & 2.31/22.3 & 1.84/22.9 & 2.21/27.3 & \textbf{2.49}/\textbf{28.9} \\
H3 (22.7\,s) & 37 & a laptop, frozen in time & 4268 (1.19\,h) & 2.15/17.6 & 1.84/18.7 & 2.24/26.1 & \textbf{2.72}/\textbf{27.4} \\
H3 (22.7\,s) & 76 & bedroom & 4408 (1.22\,h) & 2.31/20.9 & 1.87/21.8 & 2.32/26.3 & \textbf{2.84}/\textbf{27.9} \\
H3 (22.7\,s) & 75 & a motorcycle gliding through\ldots & 4391 (1.22\,h) & 2.33/21.0 & 1.88/21.1 & 2.26/25.5 & \textbf{3.38}/\textbf{26.5} \\
\midrule
Hy. (21.2\,s) & 93 & time lapse of sunrise on mars & 11702 (3.25\,h) & 2.25/24.7 & 1.88/26.9 & 2.80/32.6 & \textbf{2.91}/\textbf{38.0} \\
Hy. (21.2\,s) & 4 & a person & 11572 (3.21\,h) & 2.26/21.8 & 1.88/25.1 & 2.56/34.0 & \textbf{2.99}/\textbf{34.8} \\
Hy. (21.2\,s) & 9 & In a still frame, a stop sign & 11749 (3.26\,h) & 2.26/17.7 & 1.89/19.2 & 2.58/32.9 & \textbf{2.77}/\textbf{35.0} \\
Hy. (21.2\,s) & 89 & a black bicycle & 11709 (3.25\,h) & 2.25/20.7 & 1.88/21.2 & 2.44/33.8 & \textbf{2.80}/\textbf{34.1} \\
Hy. (21.2\,s) & 48 & A person is tasting beer & 11580 (3.22\,h) & 2.26/17.5 & 1.89/18.4 & 2.44/33.8 & \textbf{2.72}/\textbf{35.0} \\
Hy. (21.2\,s) & 90 & a parking meter on the right\ldots & 11516 (3.20\,h) & 2.26/18.9 & 1.89/20.8 & 2.26/34.5 & \textbf{2.61}/\textbf{35.5} \\
Hy. (21.2\,s) & 20 & A beautiful coastal beach in\ldots & 11545 (3.21\,h) & 2.25/20.2 & 1.88/23.2 & 2.76/29.5 & \textbf{2.94}/\textbf{32.9} \\
Hy. (21.2\,s) & 28 & Sunset time lapse at the bea\ldots & 11539 (3.21\,h) & 2.25/24.2 & 1.88/23.9 & 2.65/29.5 & \textbf{2.88}/\textbf{32.4} \\
Hy. (21.2\,s) & 39 & A beautiful coastal beach in\ldots & 11295 (3.14\,h) & 2.26/20.3 & 1.89/22.6 & 2.77/29.7 & \textbf{2.93}/\textbf{32.7} \\
Hy. (21.2\,s) & 68 & beach & 11591 (3.22\,h) & 2.26/24.4 & 1.89/29.2 & 2.60/32.0 & \textbf{2.95}/\textbf{33.0} \\
\midrule
Wan (20.8\,s) & 9 & In a still frame, a stop sign & 2173 (0.60\,h) & 2.25/23.8 & 1.89/24.2 & 2.28/33.6 & \textbf{2.68}/\textbf{36.9} \\
Wan (20.8\,s) & 94 & A person is eating watermelon & 2198 (0.61\,h) & 2.28/15.4 & 1.86/15.9 & 2.31/32.0 & \textbf{2.61}/\textbf{35.1} \\
Wan (20.8\,s) & 12 & a person eating a burger & 2384 (0.66\,h) & 2.27/16.6 & 1.91/17.1 & 2.30/32.3 & \textbf{2.36}/\textbf{36.5} \\
Wan (20.8\,s) & 37 & a laptop, frozen in time & 2232 (0.62\,h) & 2.29/17.3 & 1.94/18.0 & 2.26/32.6 & \textbf{2.73}/\textbf{34.8} \\
Wan (20.8\,s) & 93 & time lapse of sunrise on mars & 2174 (0.60\,h) & 1.98/24.9 & 1.70/25.4 & 2.42/34.8 & \textbf{2.64}/\textbf{37.6} \\
Wan (20.8\,s) & 96 & an airplane landing smoothly\ldots & 2172 (0.60\,h) & 2.19/20.0 & 1.87/20.2 & 2.20/32.8 & \textbf{2.60}/\textbf{35.7} \\
Wan (20.8\,s) & 16 & a cat and a dog & 2435 (0.68\,h) & 2.50/13.8 & 1.91/14.1 & 2.13/29.9 & \textbf{2.78}/\textbf{32.4} \\
Wan (20.8\,s) & 28 & Sunset time lapse at the bea\ldots & 2197 (0.61\,h) & 2.22/18.1 & 1.72/18.3 & 2.26/32.3 & \textbf{2.73}/\textbf{34.0} \\
Wan (20.8\,s) & 47 & ballroom & 2197 (0.61\,h) & \textbf{2.26}/20.2 & 1.86/20.7 & 2.21/33.2 & 2.25/\textbf{35.6} \\
Wan (20.8\,s) & 87 & a truck & 2524 (0.70\,h) & 2.55/19.3 & 2.19/20.5 & 2.47/32.4 & \textbf{2.73}/\textbf{33.9} \\
\bottomrule
\end{tabular}
\end{table*}

\begin{figure*}[p]
  \centering
  \includegraphics[width=\textwidth]{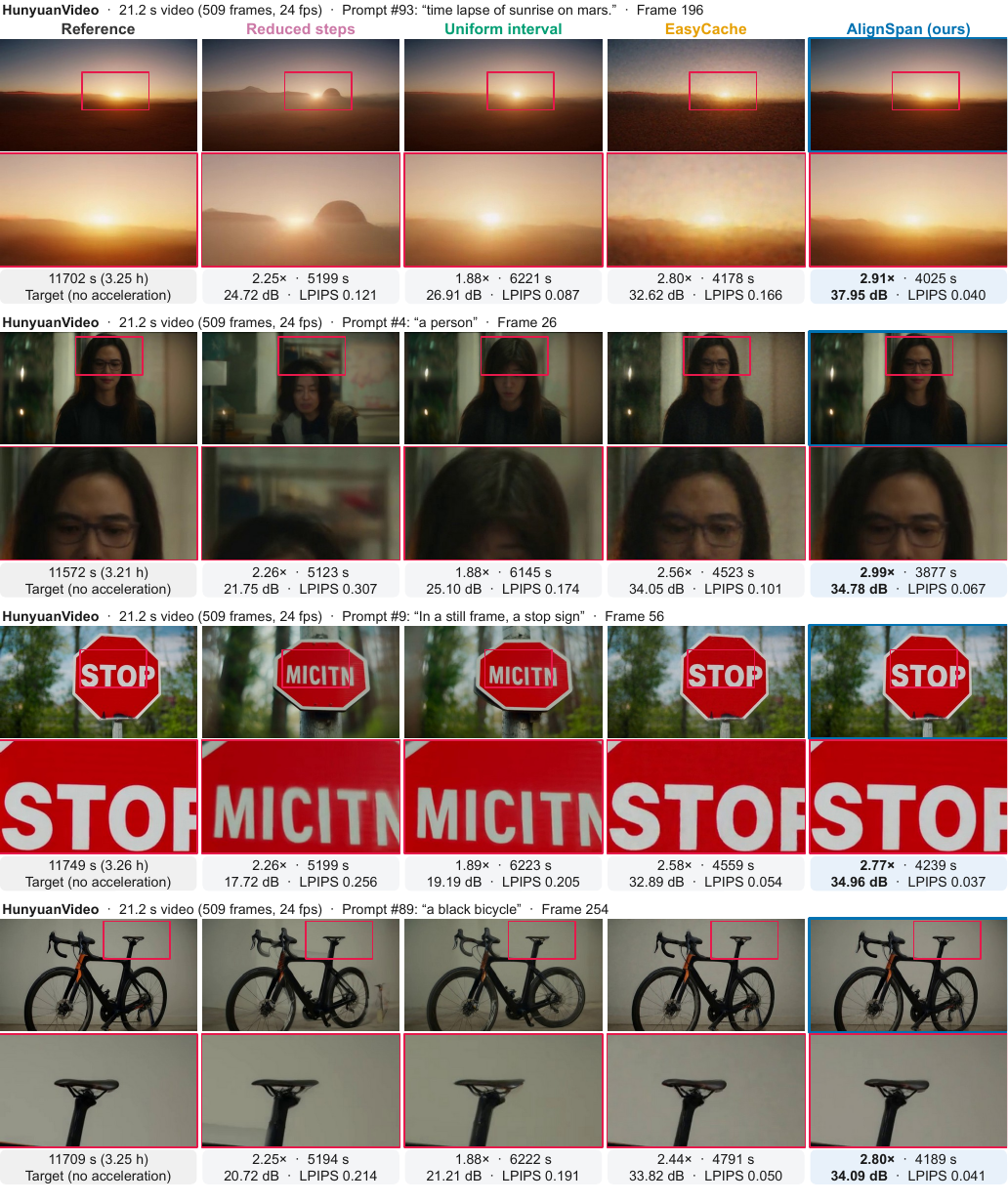}
  \caption{\textbf{Qualitative comparison, HunyuanVideo (21.2\,s videos), cases 1--4.} Each block: prompt and frame index, one frame per method with the zoom region boxed, the zoomed crop, and this video's generation time, speedup, PSNR, and LPIPS against the reference (best accelerator in bold).}
  \label{fig:qual1}
\end{figure*}

\begin{figure*}[p]
  \centering
  \includegraphics[width=\textwidth]{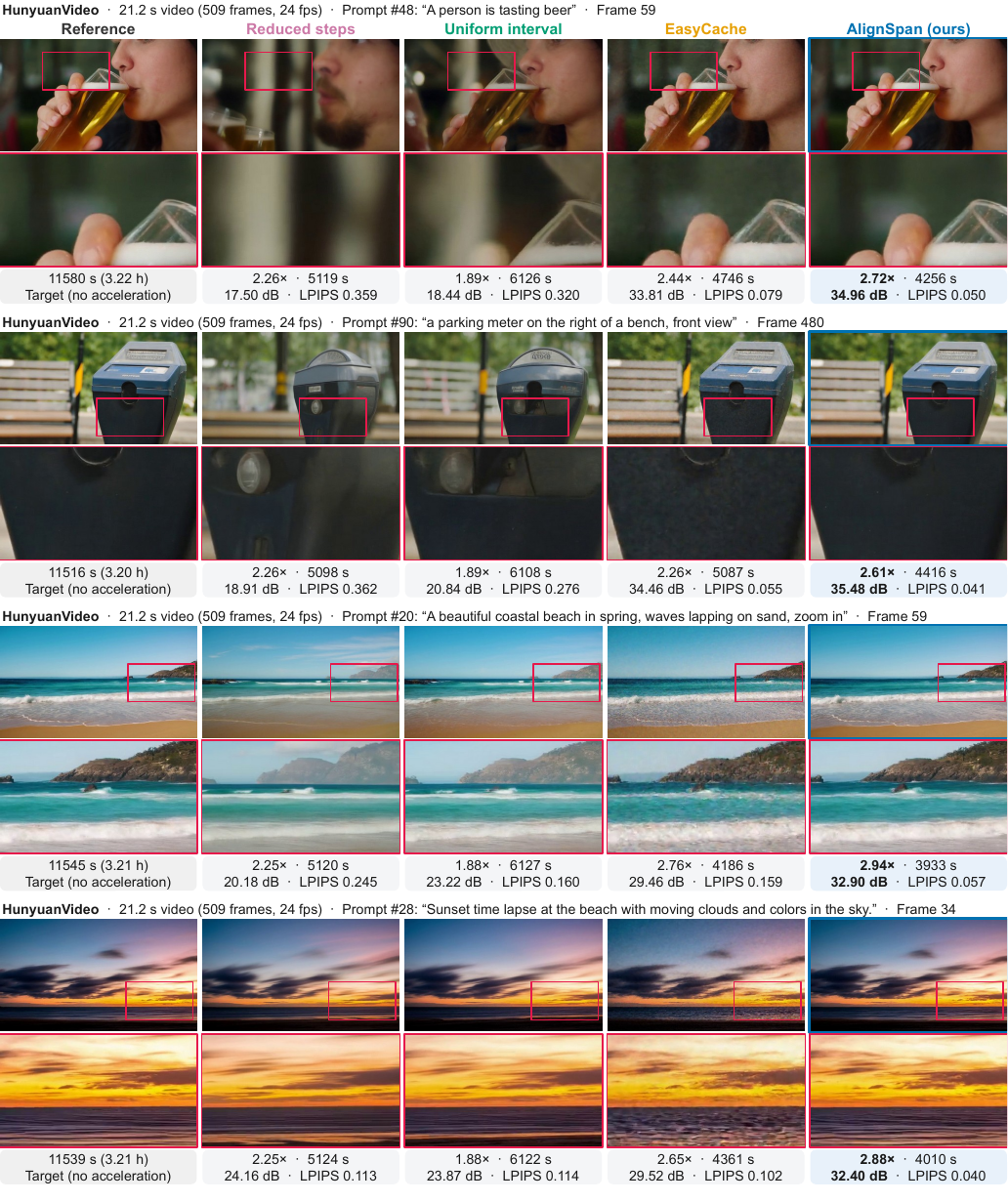}
  \caption{\textbf{Qualitative comparison, HunyuanVideo (21.2\,s videos), cases 5--8,} same layout as \cref{fig:qual1}.}
  \label{fig:qual2}
\end{figure*}

\begin{figure*}[p]
  \centering
  \includegraphics[width=\textwidth]{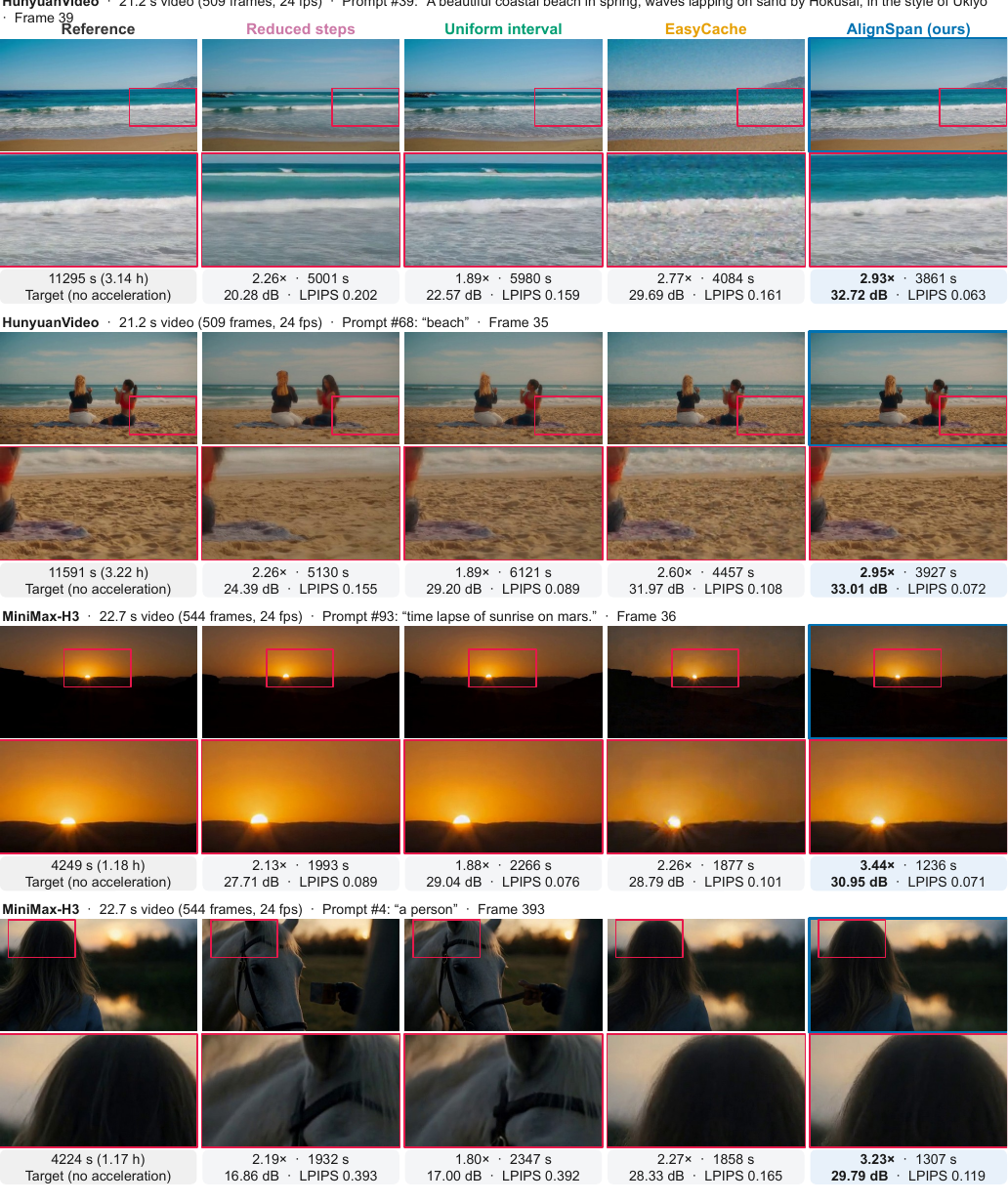}
  \caption{\textbf{Qualitative comparison, HunyuanVideo (21.2\,s) cases 9--10 and MiniMax-H3 (22.7\,s) cases 1--2,} same layout as \cref{fig:qual1}.}
  \label{fig:qual3}
\end{figure*}

\begin{figure*}[p]
  \centering
  \includegraphics[width=\textwidth]{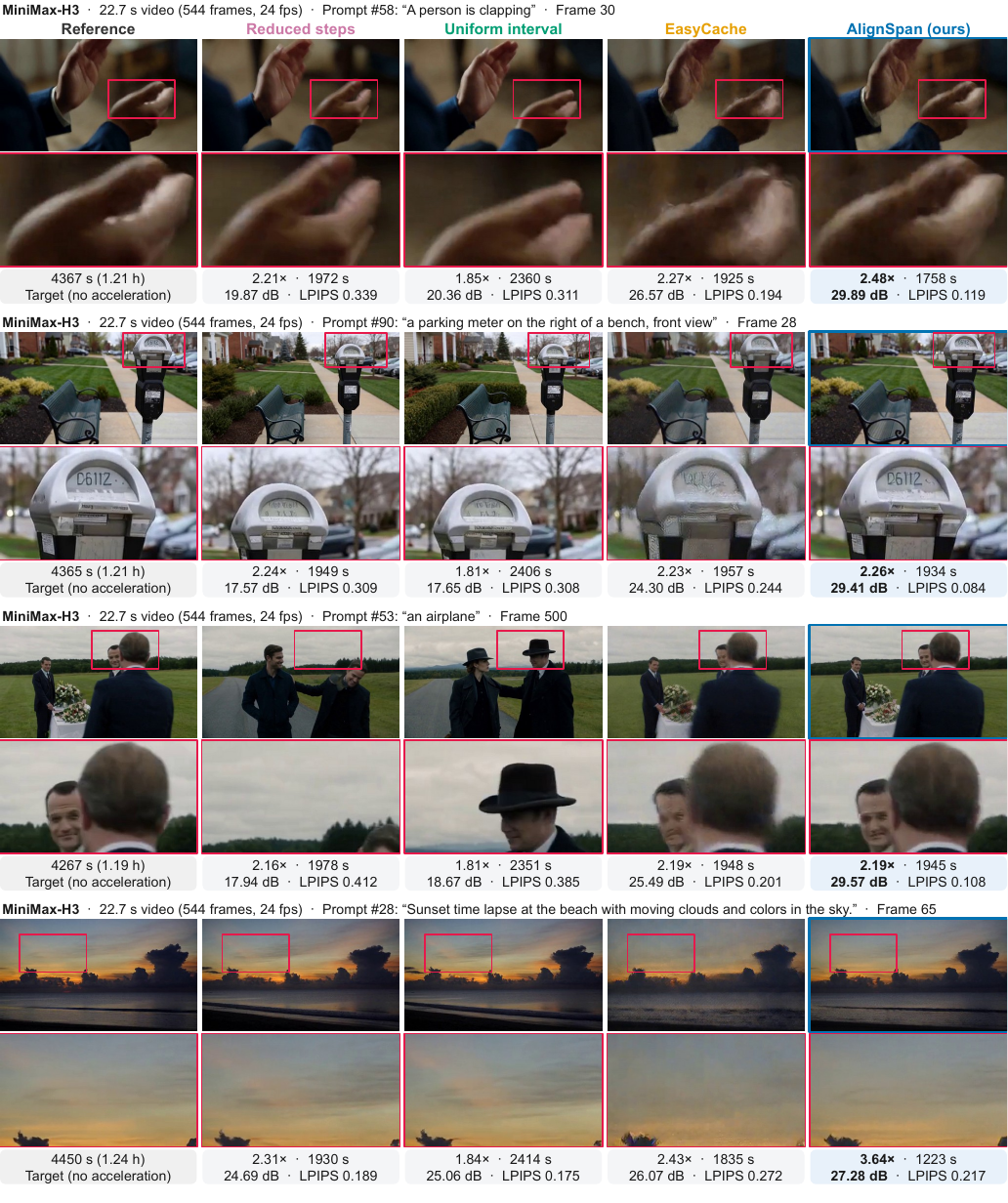}
  \caption{\textbf{Qualitative comparison, MiniMax-H3 (22.7\,s videos), cases 3--6,} same layout as \cref{fig:qual1}.}
  \label{fig:qual4}
\end{figure*}

\begin{figure*}[p]
  \centering
  \includegraphics[width=\textwidth]{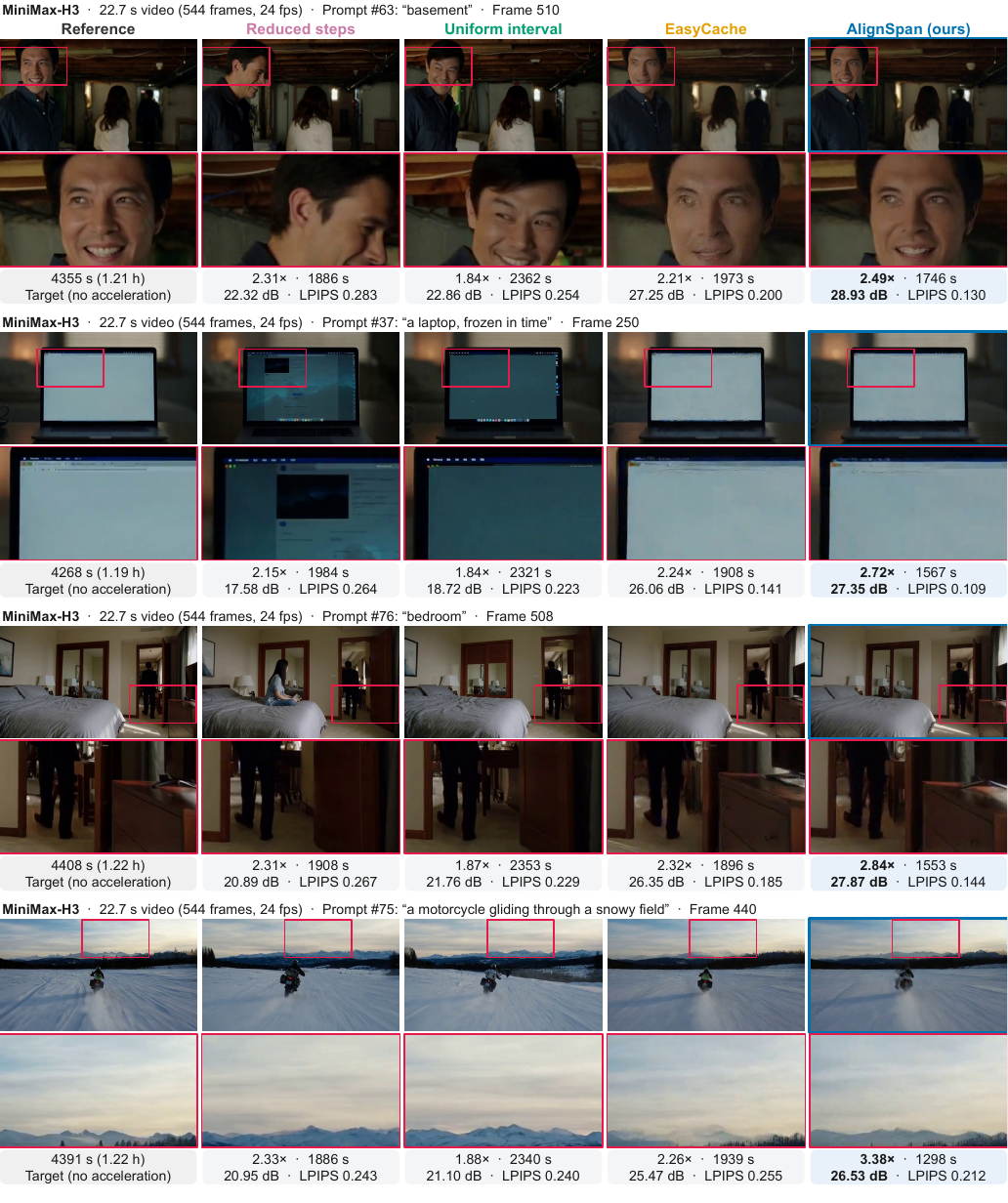}
  \caption{\textbf{Qualitative comparison, MiniMax-H3 (22.7\,s videos), cases 7--10,} same layout as \cref{fig:qual1}.}
  \label{fig:qual5}
\end{figure*}

\begin{figure*}[p]
  \centering
  \includegraphics[width=\textwidth]{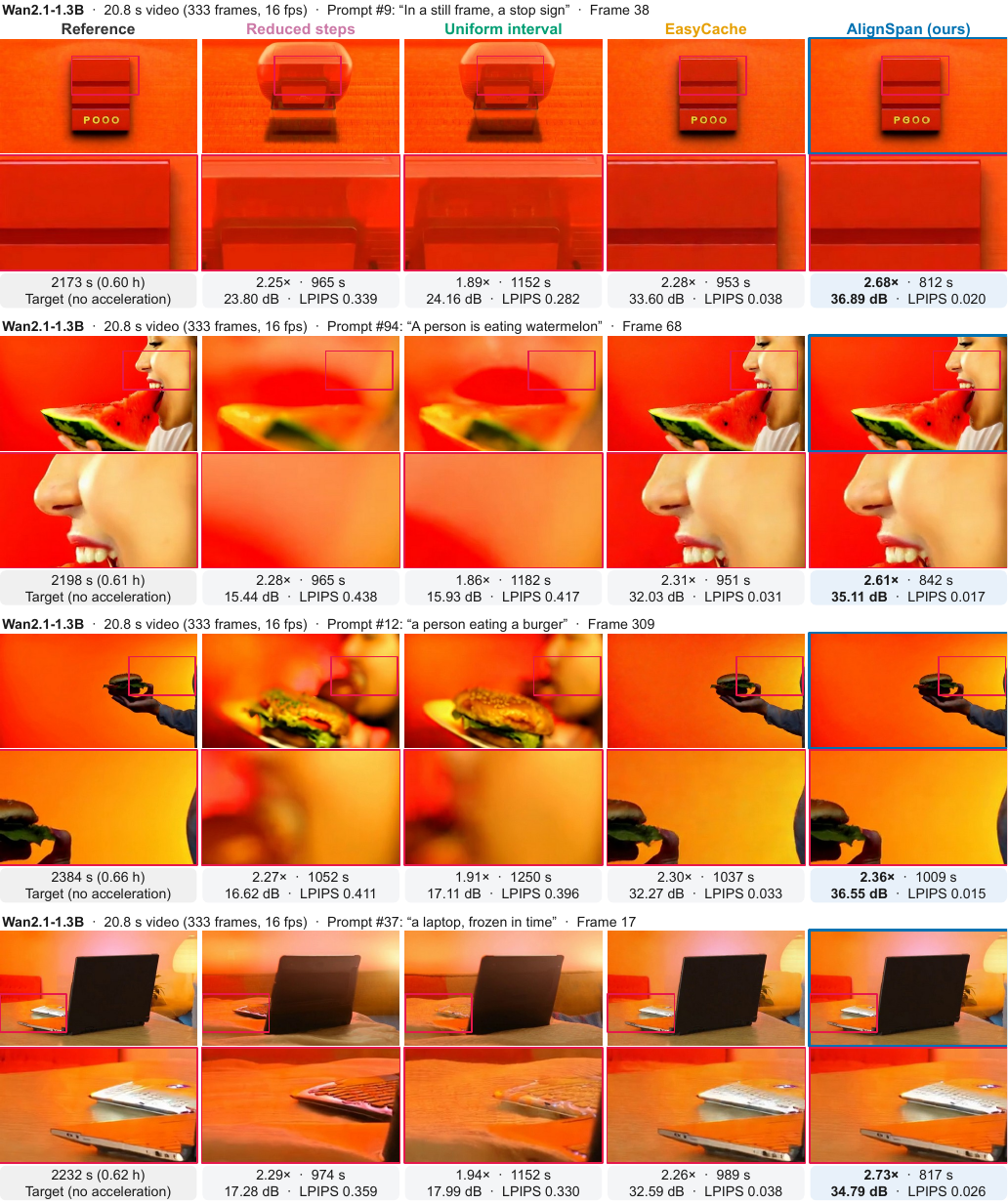}
  \caption{\textbf{Qualitative comparison, Wan2.1-1.3B (20.8\,s videos), cases 1--4,} same layout as \cref{fig:qual1}.}
  \label{fig:qual6}
\end{figure*}

\begin{figure*}[p]
  \centering
  \includegraphics[width=\textwidth]{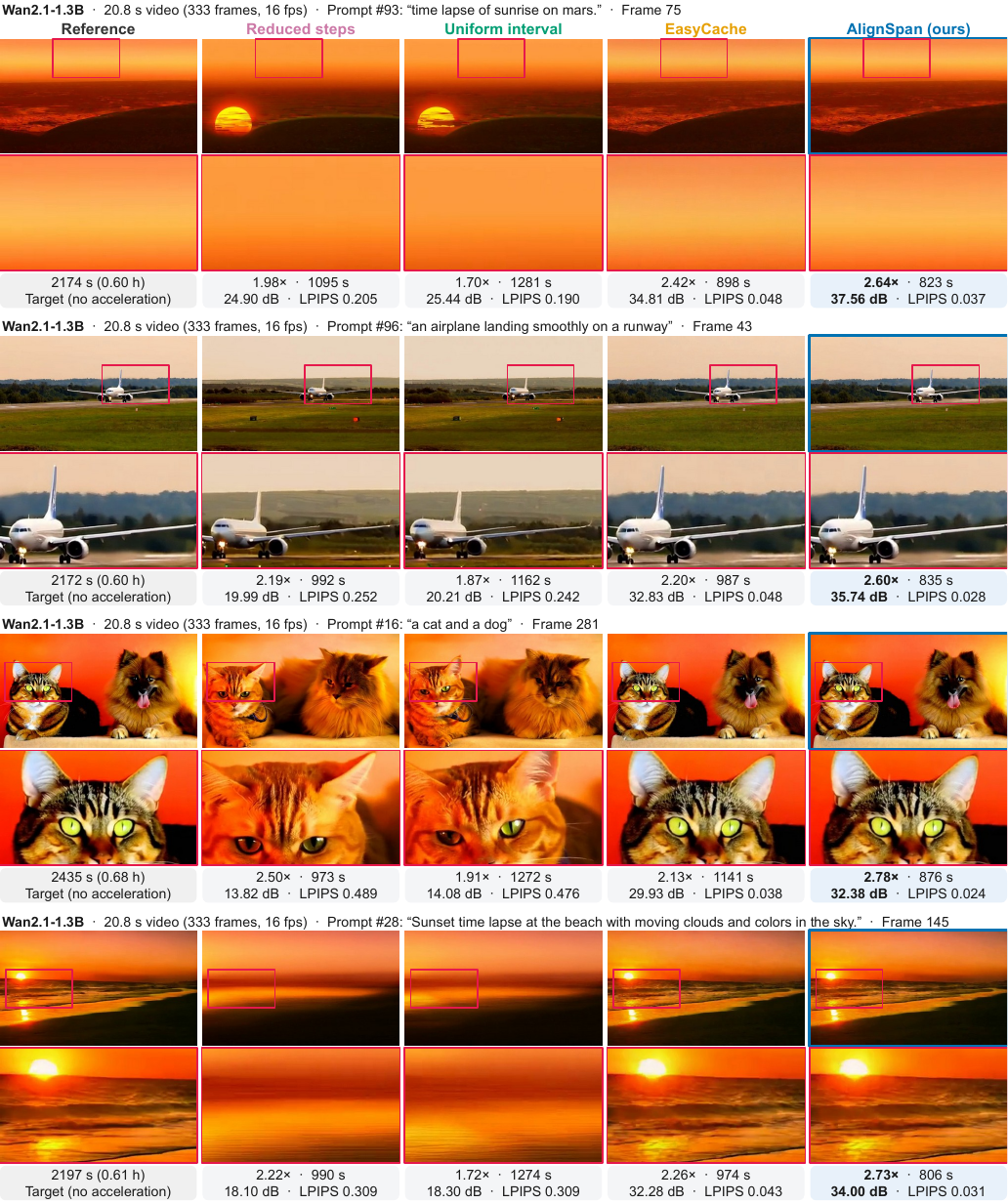}
  \caption{\textbf{Qualitative comparison, Wan2.1-1.3B (20.8\,s videos), cases 5--8,} same layout as \cref{fig:qual1}.}
  \label{fig:qual7}
\end{figure*}

\begin{figure*}[t]
  \centering
  \includegraphics[width=\textwidth]{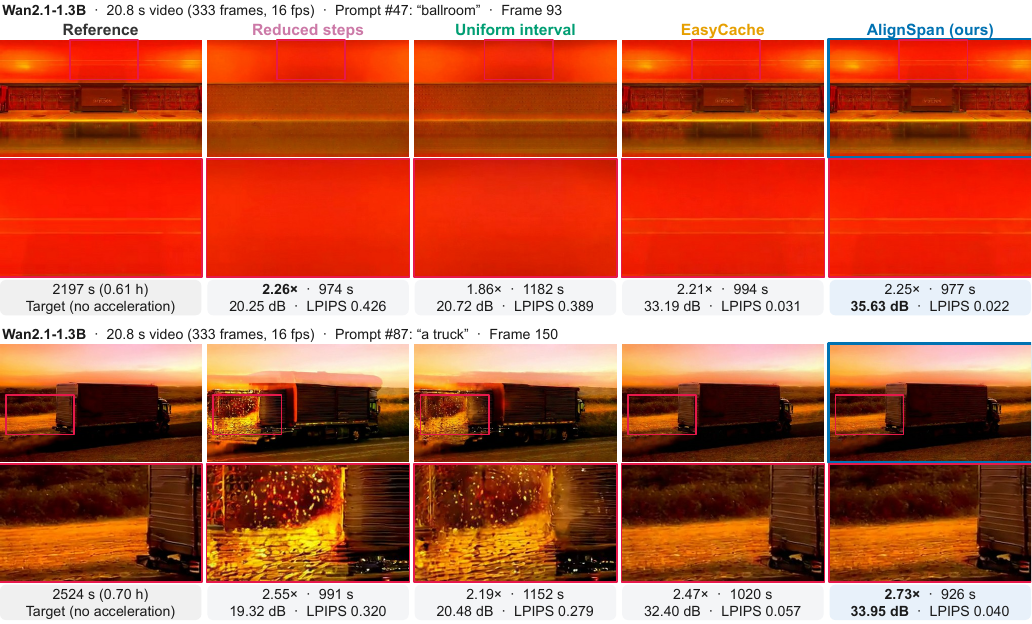}
  \caption{\textbf{Qualitative comparison, Wan2.1-1.3B (20.8\,s videos), cases 9--10,} same layout as \cref{fig:qual1}.}
  \label{fig:qual8}
\end{figure*}

\Cref{fig:qual1,fig:qual2,fig:qual3,fig:qual4,fig:qual5,fig:qual6,fig:qual7,fig:qual8} show 30 videos of the 100-prompt VBench panel (10 per backbone, $N{=}4$, 20.8--22.7\,s) with five methods side by side: one frame per method, a zoomed crop, and the generation time, speedup, and fidelity of that video.
Cases were selected automatically by the gap between \method and the other methods in a three-frame window, and the zoom region is placed where the outputs differ most from the reference or on semantically important content (text, faces, hands).
The same failure modes appear on all backbones.
Reduced steps and uniform caching often change the content: a stop sign reads ``MICITN'', a woman drinking beer becomes a bearded man, a smiling man turns away, a tabby cat becomes a ginger one.
EasyCache keeps the content but adds grain and smears fine structure such as text, hair, and fur.
\Cref{tab:cases} lists the speed and PSNR of every method for all 30 cases.
\method has the best PSNR in all 30 and the highest speed in 29, one of them (MiniMax-H3 prompt 53) tied with EasyCache; the exception is Wan2.1 prompt 47, where reduced steps is $0.01\times$ faster at $15.4$\,dB lower PSNR.

\end{document}